%% file: main.tex
\documentclass[pdflatex,sn-mathphys-num,Numbered]{sn-jnl}

\newif\ifrevealauthors
\newif\ifrevealcode
\revealauthorstrue                  
\revealcodetrue                     

\usepackage{graphicx}
\graphicspath{{./figs/}{../exp/results/figures/}{./}}
\usepackage{booktabs}
\usepackage{amsmath,amssymb}
\usepackage{algorithm}
\usepackage{algpseudocode}
\hypersetup{hidelinks}                
\usepackage{cleveref}
\usepackage{placeins}                

\usepackage{xcolor}
\usepackage{enumitem}
\usepackage{tikz}
\usetikzlibrary{positioning,arrows.meta,shapes.geometric}
\usepackage{url}

\definecolor{accentblue}{HTML}{1F77B4}   
\definecolor{accenttint}{HTML}{E6EFFA}   
\definecolor{accentwarm}{HTML}{FF7F0E}   
\definecolor{rolegreen}{HTML}{2CA02C}    
\definecolor{rolegray}{HTML}{7F7F7F}     
\definecolor{inkgray}{HTML}{333333}

\definecolor{orcidlogocol}{HTML}{A6CE39}
\newcommand{\orcidicon}{%
  \mbox{\begin{tikzpicture}[x=1ex,y=1ex,baseline={([yshift=-0.45ex]current bounding box.center)}]%
    \fill[orcidlogocol] (0,0) circle (0.95);%
    \node[white,font=\fontsize{3pt}{3pt}\selectfont\bfseries\sffamily] at (0,0) {iD};%
  \end{tikzpicture}}%
}
\newcommand{\orcidlink}[1]{\href{https://orcid.org/#1}{\orcidicon}}

\begin{document}

\title[Regression Test Selection for Updated Capability Modules]{Regression Test Selection for Updated Capability Modules in Compositional ML Systems via Atomic-Quality Probes}

\ifrevealauthors
\author[1]{\fnm{Xue} \sur{Qin}~\orcidlink{0009-0009-3642-2663}}\email{qinxue@me.com}

\author[2]{\fnm{Simin} \sur{Luan}~\orcidlink{0000-0003-1138-1892}}\email{luansiminiot@gmail.com}

\author*[3]{\fnm{Cong} \sur{Yang}~\orcidlink{0000-0002-8314-0935}}\email{cong.yang@suda.edu.cn}

\author*[2]{\fnm{Zhijun} \sur{Li}~\orcidlink{0000-0001-9129-9957}}\email{lizhijun\_os@hit.edu.cn}

\affil*[1]{\orgdiv{School of Software}, \orgname{Harbin Institute of Technology}, \orgaddress{\city{Harbin}, \country{China}}}

\affil[2]{\orgdiv{School of Computer Science and Technology}, \orgname{Harbin Institute of Technology}, \orgaddress{\city{Harbin}, \country{China}}}

\affil[3]{\orgdiv{School of Future Science and Engineering}, \orgname{Soochow University}, \orgaddress{\city{Suzhou}, \country{China}}}
\else
\author[1]{Anonymized for blind review}
\affil[1]{\orgname{Anonymized}}
\fi

\abstract{Compositional machine-learning (ML) systems assemble runtime behavior from libraries of independently re-trained capability modules. Replacing one module raises a regression-testing question that static dependence analysis cannot answer: which existing compositions stay valid, and at what test cost? We frame capability updates as regression test selection (RTS) and contribute four results. First, a paired cross-version swap protocol isolates the marginal effect of a single module update. Second, on two contact-rich manipulation tasks we characterize a \emph{dominant-skill effect}: one Embodied Capability Module (ECM) reaches $88.0\%$ atomic success while siblings stay at or below $32.0\%$, and its inclusion shifts composition success by up to $52$ percentage points; a controlled weight-space interpolation tracks composition success against atomic quality point-by-point (pooled Pearson $r{=}0.94$), and the effect replicates on a second task, where the governing module must lie on the critical path of the phase sequence. Third, off-policy behavioral-distance metrics fail to identify the dominant module. Fourth, a margin-gated Hybrid Selector matches full revalidation at zero per-decision test cost ($75.0\%$ gold-label agreement, no detectable difference) and reaches $81.25\%$ at half of full-revalidation cost, beating a cost-matched random budget (Monte-Carlo $p{=}0.039$). A resolution analysis shows that coarse evaluation overstates the apparent advantage of full revalidation. The atomic-quality probe is a principled test-selection criterion for capability-update regression testing in compositional ML systems.}

\keywords{regression testing, test selection, software reliability, ML systems testing, capability-module testing, compositional learning, statistical software testing}

\maketitle

\section{Introduction}
\label{sec:intro}

\begin{figure}[t]
\centering
\includegraphics[width=\linewidth]{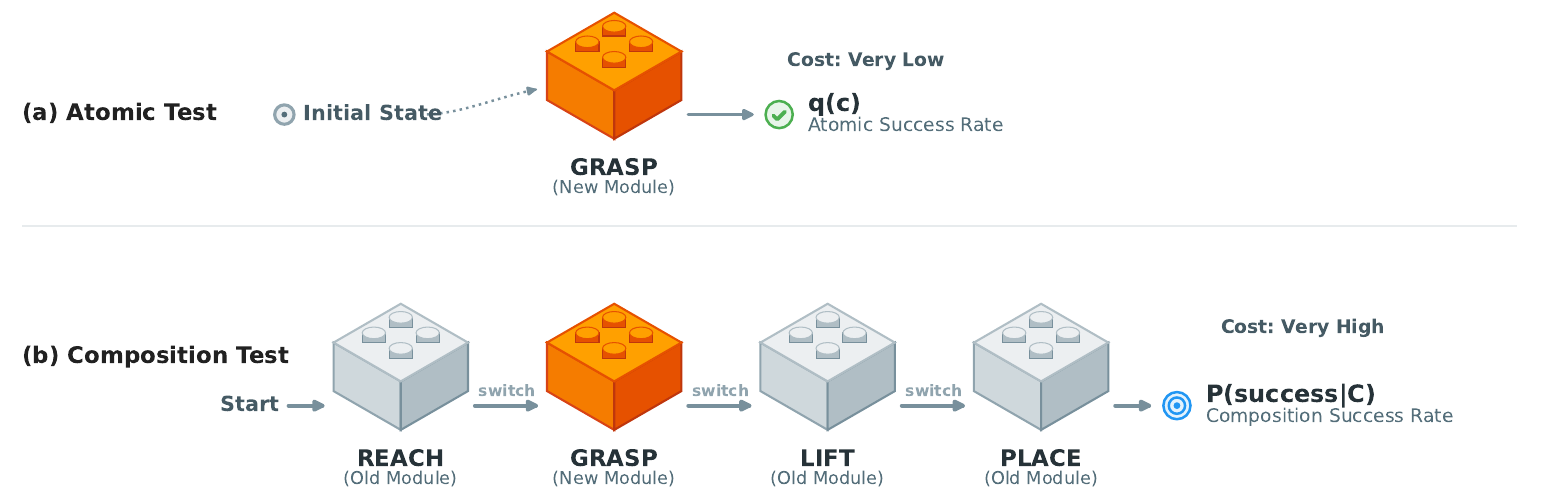}
\caption{Atomic test versus composition test for a capability-module update.}
\label{fig:concept}
\end{figure}

Compositional machine-learning (ML) systems (those that assemble runtime
behavior from a library of independently-trained capability modules)
face a software-testing problem that current methodology does not
adequately address. When one module in the library is replaced by a
re-trained or fine-tuned alternative, which of the existing
compositions that depend on it remain reliable in deployment, and at
what test cost can this be established? \Cref{fig:concept} contrasts the
two extremes of test cost for a single updated skill: a per-skill
\emph{atomic probe} that exercises the new module from its own initial
state, and a full \emph{composition test} that re-runs the entire
deployed skill chain end to end. Today this question is
answered through full revalidation across all downstream tasks. The
regression-testing literature has long catalogued this cost problem
and its mitigations under the headings of test-suite minimization,
selection, and prioritization \citep{yoo2012regression}. Those
mitigations key on source-code dependence analysis
\citep{rothermel1996analyzing}, and therefore do not transfer directly
when the ``change'' under test is a re-trained learned module whose
behavioral drift is invisible to static analysis.
\Cref{fig:hero} positions the test-selection layer this paper
develops across the seven manipulation tasks of our evaluation.

\begin{figure}[t]
\centering
\resizebox{\linewidth}{!}{%
\begin{tikzpicture}[
  every node/.style={font=\scriptsize, text=inkgray},
  task/.style={rectangle, draw=inkgray, line width=0.9pt, rounded corners=1pt,
               text width=2.70cm, minimum width=2.86cm, minimum height=1.20cm,
               align=center, inner sep=3pt, fill=white},
  govblock/.style={rectangle, draw=accentblue, line width=0.6pt, rounded corners=1pt,
                   minimum height=1.0cm, minimum width=21.63cm,
                   align=center, inner sep=5pt, fill=accenttint,
                   font=\footnotesize},
  outblock/.style={font=\footnotesize\itshape, text=inkgray!80,
                   minimum width=21.63cm, align=center},
  flow/.style={-{Latex[length=3.5pt,width=2.8pt]}, line width=0.45pt, draw=inkgray!75}
]
\node[task, draw=rolegreen] (t1) at (0.00, 0) {\textcolor{rolegreen}{\textbf{T1\_Pick}}\\[1pt]{\tiny single-arm pick}};
\node[task, draw=rolegray] (t2) at (3.09, 0) {\textcolor{rolegray}{\textbf{T2\_Place}}\\[1pt]{\tiny single-arm place}};
\node[task, draw=accentwarm] (t3) at (6.18, 0) {\textcolor{accentwarm}{\textbf{T3\_Stack}}\\[1pt]{\tiny block stacking}};
\node[task, draw=accentwarm] (t4) at (9.27, 0) {\textcolor{accentwarm}{\textbf{T4\_NutAssm}}\\[1pt]{\tiny nut assembly}};
\node[task, draw=rolegray] (t5) at (12.36,0) {\textcolor{rolegray}{\textbf{T5\_PickMulti}}\\[1pt]{\tiny multi-obj p\&p}};
\node[task, draw=accentblue] (t6) at (15.45,0) {\textcolor{accentblue}{\textbf{T6\_TwoArmPeg}}\\[1pt]{\tiny dual-arm peg-in-hole}};
\node[task, draw=accentblue] (t7) at (18.54,0) {\textcolor{accentblue}{\textbf{T7\_Door}}\\[1pt]{\tiny door opening}};
\node[govblock] (gov) at (9.27,-1.95) {{\bfseries Atomic-Probe Test-Selection Layer}\\[2pt]
  {\scriptsize atomic probe $q(c)$\;\;$\cdot$\;\;Hybrid Selector (margin $m$)\;\;$\cdot$\;\;composition probe (gold label)}};
\node[outblock] (out) at (9.27,-3.15) {\textsc{accept} / \textsc{reject}\;\;$\to$\;\;deployed phase-decomposed robot policy};
\foreach \t in {t1,t2,t3,t4,t5,t6,t7}{
  \draw[flow] (\t.south) -- (\t |- gov.north);
}
\draw[flow] (gov.south) -- (9.27,-2.80);
\end{tikzpicture}%
}
\caption{The atomic-probe test-selection criterion across the seven manipulation tasks.}
\label{fig:hero}
\end{figure}

Three lines of recent work on compositional ML systems sharpen the
gap. The typed-composition literature
\citep{liu2024blade,shao2025symskill,mishra2023gsc} treats the
capability library as fixed at test time, providing no analysis of
post-update composition reliability. The adjacent open-ended
skill-library line \citep{wang2023voyager,zhang2023boss,wan2023lotus}
\emph{grows} the library at runtime without ever studying what
happens when an existing module is \emph{replaced}. Generalist
policies designed to be fine-tuned
\citep{kim2024openvla,octo2024,black2024pi0} make post-deployment
module updates a routine event in the deployment lifecycle.

We frame this as a regression-testing problem for compositional ML
capability libraries and contribute a principled test-selection
criterion with measured safety properties. We introduce a paired-sampling cross-version swap protocol
(\cref{sec:method:swap}) that isolates the marginal effect of a single
module update while holding all other compositional context fixed.
The protocol is statistically rigorous: paired initial-state seeds
permit per-cell McNemar exact-binomial testing and a
cluster-permutation variant that respects module-level dependence
in the test population.

Applied to a standard simulation substrate
\citep{zhu2020robosuite}, the protocol reveals a behavior the
population-mean swap effect is structurally blind to. On the
representative dual-arm peg-in-hole task we discover a
\emph{dominant-skill effect}: a single ECM in the four-version candidate set achieves $88.0\%$ atomic
success while every other ECM is at or below $32.0\%$, and whether
this dominant module is included in a composition shifts the
deployment success rate by up to $+52$ percentage points
(subset-swap group mean). The single-phase paired-swap matrix on
\textsc{reach} contains gains up to $+55$pp and losses up to $-59$pp.
The same effect replicates on a second contact-rich task (Door),
where the \textsc{grasp} phase carries it: a dominant module at
$73.3\%$ atomic success lifts a failed composition by $83.3$pp, and a
dominant but terminal \textsc{place} module does not propagate,
consistent with the effect being carried by a module on the critical
path of the phase sequence.
The effect is by construction undefined on a saturated single-arm
pick task where every candidate module already achieves $100\%$,
sharply bounding the regime in which the methodology applies.
On the dual-arm task (the only task where the dominance ranking is
defined) we further find that off-policy behavioral-distance
metrics (the most natural cheap reliability predictor) fail to
identify the dominant module, ruling out the most obvious low-cost
test surrogate.

Building on these findings, we propose an \emph{atomic-quality
probe} and a \emph{Hybrid Selector} that combines a per-module
probe (zero per-decision test cost) with selective composition
revalidation (full test cost). On the dual-arm task, where a
well-defined success-rate oracle exists, the zero-cost atomic-only
probe matches the gold label on $75.0\%$ of $48$ paired update
events at $N{=}100$ evaluation resolution, with no detectable
difference from full revalidation ($75.0\%$ each; McNemar
discordant split $4{:}4$, $p{=}1.0$). The Hybrid Selector at
margin $m{=}10$ reaches
$81.25\%$, the best match of the seven selectors on this gold
label, at
$50\%$ of full-revalidation cost; it is never worse than full
revalidation on any of the $16$ update clusters, beats a
cost-matched random allocation of the same FullReval budget
(Monte-Carlo, MC, $p{=}0.039$), and matches or improves on every
alternative's unsafe rate and fault-detection sensitivity under
split-half gold labels (\cref{sec:exp:selector}). This is a Pareto-improvement on the
test-cost / test-quality frontier and a principled test-selection
criterion specifically targeting capability-update regression testing
in compositional ML systems, demonstrated here on a representative
contact-rich task.

Our contributions are:
\textbf{(i)} a paired-sampling cross-version swap protocol that
isolates per-module update effects with statistical rigor, for ML
capability libraries;
\textbf{(ii)} empirical characterization of a dominant-skill effect
and a saturation boundary, replicated on a second task (Door, $N{=}30$)
and sharpened by a critical-path condition on the dominant module's phase,
verified by a controlled weight-space
interpolation and refuting the natural off-policy
behavioral-distance surrogate as a low-cost test predictor;
\textbf{(iii)} the atomic-probe Hybrid Selector, a margin-gated
test-selection algorithm that attains the best gold-label match of
seven selectors at half of full-revalidation cost on $48$
success-rate-oracled regression-test decisions, together with an
evaluation-resolution analysis showing why coarse Monte-Carlo
evaluation overstates full revalidation; a $96$-event reward-oracle
extension is reported in \cref{sec:app:xtask}.

\input{related_work.tex}

\section{Method: A Test-Selection Criterion for Compositional Module Updates}
\label{sec:method}

We develop the atomic-quality probe as a principled test-selection
criterion for deciding whether a candidate module update should be
admitted into a compositional ML system without running the full
integration test suite. We first state the fault
model under test (\cref{sec:method:fault}), then formalize the
compositional execution model (\cref{sec:method:execution}) and the
cross-version swap protocol that serves both as our measurement
instrument and as the deployment-time abstraction of update events
(\cref{sec:method:swap}). We then introduce the central primitive,
the per-module atomic-quality probe (\cref{sec:method:atomic}), and
a family of test-selection strategies that trade per-decision cost
against oracle agreement (\cref{sec:method:selectors}).
\Cref{fig:architecture} gives the architectural overview;
\cref{sec:method:compute} states the reproducibility envelope.

\subsection{Fault Model}
\label{sec:method:fault}

We define the fault under test as follows. A \emph{composition} is a
tuple $(c_1, c_2, \ldots, c_K)$ of capability modules executed in
sequence to attempt a task. The composition's deployment
\emph{quality} is its expected success rate $\rho \in [0,1]$ under a
fixed initial-state distribution, estimated empirically by Monte
Carlo. A \emph{regression fault} is induced by replacing a single
module $c_i$ with an updated version $c_i'$ such that the
composition's success rate drops:
\begin{equation}
\rho\bigl(c_1, \ldots, c_i', \ldots, c_K\bigr) < \rho\bigl(c_1, \ldots, c_i, \ldots, c_K\bigr).
\end{equation}
A regression-test-selection criterion takes a candidate update
$(c_i, c_i')$ and the catalogue of compositions that depend on
$c_i$ and decides, per composition, whether to invoke the full
integration test (a Monte-Carlo composition rollout) or to accept
the update without testing. A criterion is \emph{safe} on a fixed
test population if it never omits a composition that would have been
flagged as faulty under full revalidation; it is
\emph{conservative} if it may include compositions that full
revalidation would have passed. Classical static-analysis
RTS~\citep{rothermel1997safe} provides safety by construction over
source-code dependencies; in our setting, where the change is a
learned-weight update, no analogous syntactic guarantee is available
and safety becomes an empirical property of the criterion. We
measure both safety (the false-negative rate, fraction of regressions
the criterion lets through) and conservatism (the false-positive
rate, fraction of non-regressions the criterion flags) explicitly in
\cref{sec:exp:safety}. One operational detail matters for
interpreting those measurements: the \emph{gold label} used as
ground truth throughout is the zero-tolerance paired comparison
(an update event is labeled a regression unless the post-update
composition's success-rate estimate on the paired episode pool
strictly exceeds the incumbent's; exact ties on the finite
evaluation pool are labeled conservatively as regressions), while
every selector, including full revalidation itself, operates at a
deployment tolerance $\tau$ ($\tau{=}5$pp on success-rate-oracled
tasks; the reward-oracle analogue is defined in
\cref{sec:exp:tasks}). The cost of operating a $\tau$-tolerant
rule against the zero-tolerance gold label is measured explicitly
in \cref{sec:exp:safety}.

\subsection{Compositional Skill Execution}
\label{sec:method:execution}

We represent a long-horizon manipulation task as an ordered sequence
of phases $\Pi = (\pi_1, \dots, \pi_K)$, with each phase served by a
phase-specific neural controller we call an \emph{Embodied Capability
Module} (ECM), following the abstraction introduced by
\citet{aeros2026}. An ECM $c$ is a temporally-abstract
option~\citep{sutton1999options}: a deterministic policy
$\pi_c: \mathcal{S} \to \mathcal{A}$ over the robot state together
with a phase-termination predicate
$\beta_c: \mathcal{S} \to \{0,1\}$. At runtime, $c_k$ is invoked on
entry to phase $\pi_k$ and is replaced by $c_{k+1}$ once $\beta_{c_k}$
fires. We instantiate ECMs as feed-forward policies trained with
Soft Actor-Critic (SAC)~\citep{haarnoja2018sac} on the robosuite
manipulation suite~\citep{zhu2020robosuite}, but the framework is
agnostic to the underlying controller class.

We use a fixed $K{=}4$ phase decomposition $\Pi = (\textsc{reach},
\textsc{grasp}, \textsc{lift}, \textsc{place})$ throughout. BLADE
\citep{liu2024blade} and SymSkill~\citep{shao2025symskill} extract
task-specific phase schedules from demonstrations or LLM prompts;
the fixed-$K$ choice trades that flexibility for a controlled
substrate on which the swap protocol below can be applied
exhaustively.

A \emph{composition} is an assignment $C = (c_1, \dots, c_K)$ with
each $c_k$ drawn from a pool of candidate ECMs $\mathcal{C}_k$
trained for phase $\pi_k$. The composition success rate
$\mathbb{P}(\mathrm{success} \mid C)$ is the probability that the
phase-scheduled execution achieves the task-level success criterion
on a paired-sampling rollout (\cref{sec:method:atomic}).

\subsection{Skill Versions and the Cross-Version Swap Protocol}
\label{sec:method:swap}

To produce multiple versions of each phase ECM we train $S{=}4$
independent SAC policies per phase, varying only the random seed
used for environment initialization, network weights, and
replay-buffer sampling. We use seeds $\{42, 7, 123, 2024\}$
throughout. This protocol simulates the realistic deployment in
which the same target skill is independently re-trained from
different demonstration batches, fine-tuning runs, or
domain-adaptation cycles, producing functionally similar but not
identical ECM populations.

For each phase $\pi_k$ we obtain a candidate set
$\mathcal{C}_k = \{c_k^{(1)}, \dots, c_k^{(S)}\}$. A
\emph{swap-set} $\sigma \subseteq \Pi$ identifies the phases whose
ECM is swapped from a \emph{primary} version $p$ to an
\emph{alternative} $a$:
\begin{equation}
C(p, a, \sigma) = \big(c_k^{(a)} \text{ if } \pi_k \in \sigma,\;
                       c_k^{(p)} \text{ otherwise}\big)_{k=1}^{K}.
\end{equation}
The diagonal cells $\sigma = \emptyset$ and $\sigma = \Pi$ are
within-version baselines; the remaining $2^K - 2$ subsets
characterize partial cross-version mixing. We use \emph{paired
episode initial states}: the same $N$ environment seeds for every
$(p, a, \sigma)$ ($N{=}100$, seeds $[10000, 10099]$, on the T6
experiments; $N{=}30$ on the reward-oracled extension tasks),
enabling paired
$t$-tests on $\Delta\mathrm{success}$ and McNemar's exact-binomial
test over the same episode pool.

The protocol serves a dual purpose. As a measurement instrument it
exposes the conditional structure of swap effects across phases
(\cref{sec:exp:mechanism}). As a deployment abstraction it
instantiates a \emph{skill-update event} $(p, a, \pi_k)$ on which
our test-selection strategies (\cref{sec:method:selectors}) operate.

\subsection{The Atomic-Quality Probe}
\label{sec:method:atomic}

The central primitive of our test-selection framework is the
\emph{atomic-quality probe} $q$, defined per ECM as
\begin{equation}
q(c) \,:=\, \mathbb{P}\!\big(\mathrm{success} \,\big|\, c \text{ controls the entire episode}\big),
\end{equation}
i.e., the probability that $c$ achieves task success when invoked as
the sole controller of an episode (regardless of phase boundaries).
The probe yields a per-ECM scalar that is reusable across every
swap-set evaluation involving $c$: with $|\mathcal{C}_k| = S$ ECMs
per phase, $KS$ atomic probes amortize across $O(S^K)$ candidate
compositions, so the probe's per-decision cost is effectively zero
at deployment.

We use binary task success rather than reward for both the atomic
probe and the composition probe defined below. Reward-based metrics
are shaping-dependent and, as our preliminary experiments confirmed,
can produce qualitatively misleading rankings on tasks where atomic
policies do not actually succeed. The atomic probe is, in spirit, a
capability-targeted behavioral test of the candidate skill in
isolation, in the lineage of probe-as-test methodologies popularized
in NLP by CheckList~\citep{ribeiro2020checklist}.

The companion expensive primitive is the \emph{composition probe}
$\mathbb{P}(\mathrm{success} \mid C)$, which runs the full
phase-scheduled composition for $N$ paired episodes per
cell. The composition probe is the expensive primitive whose cost
the test-selection framework aims to amortize.

\subsection{Test-Selection Strategies and the Hybrid Selector}
\label{sec:method:selectors}

A \emph{skill-update event} is a tuple $(p, a, \pi_k)$ specifying
that the phase-$\pi_k$ ECM in primary version $p$ is a candidate to
be replaced by the phase-$\pi_k$ ECM from version $a$. A
\emph{selector} is a function
$\mathcal{S}: (p, a, \pi_k) \to \{\textsc{accept}, \textsc{reject}\}$
that decides whether to admit the update. We compare each selector
against the \emph{gold label} computed post hoc from the paired
composition probe: an update event is accepted iff the post-update
composition's success-rate estimate on the paired episode pool
strictly exceeds the incumbent's,
$\mathbb{P}(\mathrm{success} \mid C') >
\mathbb{P}(\mathrm{success} \mid C)$, with exact ties labeled
conservatively as regressions (\cref{sec:method:fault}). Selectors
operate at a tolerance threshold $\tau$ (we use $\tau{=}5$pp
throughout). \emph{Oracle match} is the fraction of update events
where a selector's decision agrees with the gold label.

We benchmark seven selectors at progressively higher cost. The two
naive baselines are \textbf{Naive} (always accept) and
\textbf{Freeze} (always reject). The latter is the de-facto
deployment under BLADE and SymSkill, which treat the library as
immutable. \textbf{AtomicOnly} accepts iff
$q(c_a) \geq q(c_p) - \tau$, using the atomic probe per ECM at zero
per-decision cost amortized across events involving the same ECM.
\textbf{FullReval} runs the composition probe for every event and
applies the $\tau$-tolerant acceptance rule
\begin{equation}
\mathbb{P}(\mathrm{success} \mid C') \geq \mathbb{P}(\mathrm{success} \mid C) - \tau,
\end{equation}
paying one composition probe ($N$ rollouts) per event; this is
the strongest $\tau$-tolerant selector available on the same
observable. The
\textbf{Hybrid Selector}, our proposed test-selection algorithm,
trusts the cheap atomic probe when the atomic margin is large and
falls back to the composition probe only when the margin is small:

\begin{algorithm}[h]
\caption{Hybrid Selector with margin $m$.}
\label{alg:hybrid}
\begin{algorithmic}[1]
\Require Skill-update event $(p, a, \pi_k)$; atomic probes $q(\cdot)$; margin $m$; tolerance $\tau$.
\Ensure Decision $d \in \{\textsc{accept}, \textsc{reject}\}$.
\If{$\big|q(c_a^{(\pi_k)}) - q(c_p^{(\pi_k)})\big| \geq m$}
  \State \Return \textsc{accept} if $q(c_a^{(\pi_k)}) \geq q(c_p^{(\pi_k)}) - \tau$ else \textsc{reject}.
\Else
  \State Run composition probe on $C' = C(p, a, \{\pi_k\})$.
  \State \Return \textsc{accept} if $\mathbb{P}(\mathrm{success} \mid C') \geq \mathbb{P}(\mathrm{success} \mid C) - \tau$ else \textsc{reject}.
\EndIf
\end{algorithmic}
\end{algorithm}

The margin $m$ controls the cost--quality trade-off. At $m{=}0$ the
Hybrid Selector reduces to AtomicOnly; as $m \to \infty$ it reduces
to FullReval. We report $m \in \{10, 20, 30\}$pp. The expected
per-decision cost is
$N \cdot \mathbb{P}(|q(c_a) - q(c_p)| < m)$ rollouts, which
\cref{sec:exp:selector} shows is substantially below $N$ in
realistic candidate populations.

\subsection{Architecture Overview}
\label{sec:method:arch}

\Cref{fig:architecture} positions the three layers of the framework.
A robot policy is a composition of phase ECMs. A skill-update event
proposes replacing one phase's ECM with an independently-trained
alternative, the unit of change that current typed-composition
methods leave unspecified. The atomic-probe test-selection layer decides
whether to admit the update using the per-skill atomic-quality probe,
the Hybrid Selector, and selective use of the composition probe.

\begin{figure}[t]
\centering
\resizebox{\linewidth}{!}{%
\begin{tikzpicture}[
  node distance=0.55cm and 0.8cm,
  every node/.style={font=\footnotesize, text=inkgray},
  ghost/.style={rectangle, draw=inkgray, line width=0.4pt, rounded corners=1pt,
                minimum height=0.95cm, minimum width=13.8cm,
                text width=12.6cm,
                align=center, inner sep=4pt, fill=white},
  accent/.style={rectangle, draw=accentblue, line width=0.6pt, rounded corners=1pt,
                 minimum height=1.25cm, minimum width=13.8cm,
                 text width=12.6cm,
                 align=center, inner sep=4pt, fill=accenttint},
  edgelbl/.style={font=\scriptsize\itshape, text=inkgray!75,
                  midway, right=3pt, text width=4cm, align=left},
  flow/.style={-{Latex[length=4pt,width=3pt]}, line width=0.5pt, draw=inkgray}
]
\node[ghost] (policy) {{\bfseries Robot Policy}\quad{\footnotesize composition $C=(c_1,\ldots,c_K)$ of phase ECMs}};
\node[ghost, below=of policy] (update) {{\bfseries Skill Update Event}\quad $(p,a,\pi_k)$ \;\;{\footnotesize replace primary $c_p$ with alternative $c_a$ at phase $\pi_k$}};
\node[accent, below=of update] (gov) {{\bfseries Atomic-Probe Test-Selection Layer}\\[2pt]
  {\scriptsize atomic probe $q(c)$\;\;$\cdot$\;\;Hybrid Selector (margin $m$)\;\;$\cdot$\;\;composition probe (gold label)}};
\node[ghost, below=of gov] (updated) {{\bfseries Updated Robot Policy}\quad{\footnotesize composition with accepted updates}};

\draw[flow] (policy) -- node[edgelbl]{candidate update event} (update);
\draw[flow] (update) -- node[edgelbl]{screened by} (gov);
\draw[flow] (gov)    -- node[edgelbl]{\textsc{accept} / \textsc{reject}} (updated);
\end{tikzpicture}%
}
\caption{The atomic-probe test-selection pipeline.}
\label{fig:architecture}
\end{figure}

\subsection{Compute and Reproducibility}
\label{sec:method:compute}

All T1--T6 policy training ran on a single NVIDIA RTX 5090 GPU with
robosuite~1.5.2 (MuJoCo~3.7.0) as the simulator, Gymnasium~0.29.1
environment interfaces, and Soft Actor-Critic (SAC) from
Stable-Baselines3~2.8.0 on PyTorch, with EGL off-screen rendering
(\texttt{MUJOCO\_GL=egl}). The SAC training schedule is $50\mathrm{K}$
environment steps $\times 20$ iterations per phase on T3, T4, and
T6, and $50\mathrm{K} \times 15$ on T1; full hyperparameters are in
the released \texttt{configs/default.yaml}, and the exact pinned
package versions for every environment are in the released
\texttt{requirements} files. Per-seed wall-time
ranges from $2.9$\,h (T1) to $6.3$\,h (T3/T4); the four-seed
multi-task suite (T1, T3, T4 each $\times 4$ seeds) totals
$\sim 60$ GPU-hours, plus T6 checkpoints at the same four seeds
trained under the same schedule ($\sim 24$ GPU-hours each,
$\sim 96$\,h total). The second task, \texttt{T7\_Door}, was trained
separately on a cloud NVIDIA RTX GPU (CUDA~13.0, PyTorch~2.12.0)
under the identical robosuite~1.5.2 / MuJoCo~3.7.0 / SAC stack across
four phases and four seeds; to bound the accumulation of simulator
worker processes we executed one training iteration per operating-system
process, to iteration~4, and evaluated at $N{=}30$ paired episodes
(seeds $[10000, 10029]$). All T6 evaluation rollouts reported in this
paper were executed at $N{=}100$ paired episodes per cell (paired
initial-state seeds $[10000, 10099]$, $\sim 16{,}000$ episodes in
total) on an Apple M1 Max with CPU-only policy inference; the
reward-oracled extension tasks retain $N{=}30$ (seeds
$[10000, 10029]$). A fixed-prefix $N{=}30$ subsample of the
$N{=}100$ run reproduces the originally published GPU-workstation
values within sampling error (maximum deviation $8.7$pp on the
$16$-cell atomic matrix), and \cref{sec:exp:selector} reports how
conclusions depend on the evaluation resolution. The paired
structure enables the McNemar
and paired-$t$ tests reported throughout. Random seeds
$\{42, 7, 123, 2024\}$ are used for ECM training. Code and evaluation
data are available at the repositories listed in the Data
Availability statement;
additional figure-regeneration scripts and the full per-run logs
will be added as a post-acceptance release.

\section{Experiments}
\label{sec:experiments}

We evaluate the atomic-probe test-selection criterion along four lines: a positive
demonstration of the dominant-skill effect on a contact-rich
peg-in-hole task, replicated on a second contact-rich task
(\cref{sec:exp:mechanism,sec:exp:door}); a saturation boundary
on a single-arm pick task where the effect is by construction
undefined (\cref{sec:exp:boundary}); a refutation of off-policy
behavioral distance as a cheap predictor of the dominant ECM
(\cref{sec:exp:bdist}); and a cost--quality benchmark of the Hybrid
Selector, success-rate-oracled on T6 and extended to T3 and T4
under a reward oracle (\cref{sec:exp:selector}).

\subsection{Tasks, Training Setup, and Evaluation Protocol}
\label{sec:exp:tasks}

We evaluate on seven tasks from the robosuite manipulation
benchmark~\citep{zhu2020robosuite} (\cref{fig:task_panel}). Six form
a systematic sweep across manipulation skill types: single-arm pick
and place (\textsc{t1}--\textsc{t2}), stacking and nut assembly
(\textsc{t3}--\textsc{t4}), multi-object pick-and-place
(\textsc{t5}), and dual-arm peg-in-hole (\textsc{t6}). The seventh,
contact-rich door opening (\textsc{t7}), was added to test whether
the dominant-skill effect of \cref{sec:exp:mechanism} generalizes
(\cref{sec:exp:door}). We use the
task identifiers \texttt{T1\_Pick}, \texttt{T2\_Place},
\texttt{T3\_Stack}, \texttt{T4\_NutAssembly},
\texttt{T5\_PickPlaceMulti}, \texttt{T6\_TwoArmPegInHole}, and
\texttt{T7\_Door}
throughout. Each task is decomposed into the same four phases
$\Pi = (\textsc{reach}, \textsc{grasp}, \textsc{lift},
\textsc{place})$ defined in \cref{sec:method:execution}, with
phase-specific ECMs trained from scratch using SAC under the
schedule of \cref{sec:method:compute}.

\begin{figure}[t]
\centering
\resizebox{\linewidth}{!}{%
\begin{tikzpicture}[
  every node/.style={font=\footnotesize, text=inkgray},
  task/.style={rectangle, draw=inkgray, line width=0.9pt, rounded corners=1pt,
               minimum width=4.1cm, minimum height=2.20cm,
               inner sep=5pt, fill=white}
]
\node[task, draw=rolegreen] at (0.00, 0.00) {%
  \begin{minipage}[c][1.85cm][s]{3.55cm}\centering
    {\textcolor{rolegreen}{\bfseries T1\_Pick}}\\[2pt]
    {\scriptsize single-arm pick of one object}
    \vfill
    {\scriptsize\itshape\color{rolegreen}boundary case~(\cref{sec:exp:boundary})}
  \end{minipage}};
\node[task, draw=rolegray] at (4.55, 0.00) {%
  \begin{minipage}[c][1.85cm][s]{3.55cm}\centering
    {\textcolor{rolegray}{\bfseries T2\_Place}}\\[2pt]
    {\scriptsize single-arm place onto target}
    \vfill
    {\scriptsize\itshape\color{rolegray}$0\%$ atomic\;\;$\cdot$\;\;out of scope}
  \end{minipage}};
\node[task, draw=accentwarm] at (9.10, 0.00) {%
  \begin{minipage}[c][1.85cm][s]{3.55cm}\centering
    {\textcolor{accentwarm}{\bfseries T3\_Stack}}\\[2pt]
    {\scriptsize three-block stacking}
    \vfill
    {\scriptsize\itshape\color{accentwarm}reward-oracle ext.~(\cref{sec:app:xtask})}
  \end{minipage}};
\node[task, draw=accentwarm] at (0.00,-2.55) {%
  \begin{minipage}[c][1.85cm][s]{3.55cm}\centering
    {\textcolor{accentwarm}{\bfseries T4\_NutAssembly}}\\[2pt]
    {\scriptsize nut placement on peg}
    \vfill
    {\scriptsize\itshape\color{accentwarm}reward-oracle ext.~(\cref{sec:app:xtask})}
  \end{minipage}};
\node[task, draw=rolegray] at (4.55,-2.55) {%
  \begin{minipage}[c][1.85cm][s]{3.55cm}\centering
    {\textcolor{rolegray}{\bfseries T5\_PickPlaceMulti}}\\[2pt]
    {\scriptsize multi-object pick \& place}
    \vfill
    {\scriptsize\itshape\color{rolegray}$0\%$ atomic\;\;$\cdot$\;\;out of scope}
  \end{minipage}};
\node[task, draw=accentblue] at (9.10,-2.55) {%
  \begin{minipage}[c][1.85cm][s]{3.55cm}\centering
    {\textcolor{accentblue}{\bfseries T6\_TwoArmPegInHole}}\\[2pt]
    {\scriptsize dual-arm contact-rich peg insertion}
    \vfill
    {\scriptsize\itshape\color{accentblue}dominant-skill~(\cref{sec:exp:mechanism})}
  \end{minipage}};
\node[task, draw=accentblue] at (4.55,-5.10) {%
  \begin{minipage}[c][1.85cm][s]{3.55cm}\centering
    {\textcolor{accentblue}{\bfseries T7\_Door}}\\[2pt]
    {\scriptsize contact-rich door opening}
    \vfill
    {\scriptsize\itshape\color{accentblue}dominant-skill replication~(\cref{sec:exp:door})}
  \end{minipage}};
\end{tikzpicture}%
}
\caption{The seven robosuite manipulation tasks and their evaluation roles.}
\label{fig:task_panel}
\end{figure}

\subsubsection{Where each effect is observable}
The dominant-skill effect (\cref{sec:exp:mechanism}) requires (a)
atomic ECMs that achieve non-trivial task success and (b) atomic
quality that varies across versions. Among the six tasks of our
initial sweep this holds
only for T6\_TwoArmPegInHole; on T1\_Pick all atomic ECMs saturate at
$100\%$ (boundary case, \cref{sec:exp:boundary}); on T2--T5 all
atomic ECMs score $0\%$ under our standard schedule, so the effect
is undefined there (\cref{sec:lim}). We use T6 for the dominant-skill
effect and T1 for the boundary, and the added T7\_Door, on which the
effect replicates, provides a second positive task
(\cref{sec:exp:door}). For the behavioral-distance
measurements (\cref{sec:exp:bdist}) and the reward-oracle extension
of the selector benchmark
(\cref{sec:exp:selector}) we additionally use T3 and T4, on which
the oracle is necessarily defined on episode reward rather than task
success, a mixed-oracle caveat we surface explicitly at every use.
Concretely, on T3 and T4 every quantity substitutes the mean
episode return over the same $N{=}30$ paired initial states for the
success rate: the gold label accepts iff the post-update
composition's mean return strictly exceeds the incumbent's, and the
selectors apply the same numeric threshold $\tau{=}5$ on the raw
reward scale. Because this tolerance is not scale-matched to the
success-rate tolerance, reward-oracled and SR-oracled match rates
are not commensurable, which is one more reason the cross-task
aggregate (\cref{sec:app:xtask}) is reported as exploratory rather
than primary evidence.

\subsubsection{Evaluation protocol}
All experiments use the paired cross-version swap protocol of
\cref{sec:method:swap}: $S{=}4$ independently-trained ECMs per phase
under seeds $\{42, 7, 123, 2024\}$, $N{=}100$ paired episodes per
cell on T6 (init-state seeds $[10000, 10099]$), $N{=}30$ on the
replication task T7\_Door, and $N{=}30$ on the
reward-oracled T3 and T4. Statistical claims
report bootstrap $95\%$ CIs ($B{=}5000$) following
\citet{agarwal2021precipice}; pairwise comparisons use McNemar's
exact-binomial test, with a cluster-permutation variant where event
dependence is structural (\cref{sec:exp:selector}).

\subsection{The Dominant-Skill Effect}
\label{sec:exp:mechanism}

A naive expectation is that swapping one phase's ECM for an
independently-trained sibling will perturb composition outcomes one
way or the other. At the level of population mean this expectation
fails precisely where the effect is largest: on \textsc{reach},
the phase that carries gains up to $+55$pp and losses up to
$-59$pp, the paired $t$-test on $\Delta\mathrm{success}$ over the
full $4 \times 4$ cross-seed swap matrix gives $p{=}0.90$, because
gains toward and losses away from the dominant module cancel in
expectation; \textsc{lift} and \textsc{place} are likewise
unrejected ($p{=}0.22$ and $0.42$). \textsc{grasp} shows a small
positive mean shift ($+5.7$pp, Holm-adjusted $p{=}0.007$), an
order of magnitude smaller than the conditional \textsc{reach}
effects. The
cell-level variance of composition success rate \emph{inflates}
under swap by $1.09\times$ to $1.37\times$ across the four phases
relative to the within-version diagonal, indicating structured
rather than random perturbation. We show below that some swaps help
while others hurt in a way that cancels in expectation:
\emph{one specific ECM in the candidate set is disproportionately
responsible for composition success, and the sign of any swap is
determined by whether this dominant ECM enters or leaves the
composition.} All proportions in this paper are reported with
$95\%$ bootstrap confidence intervals (CIs; B${=}5000$ resamples)
where space permits; full per-cell intervals are tabulated in
\cref{sec:app:ci}.

\subsubsection{Atomic quality is concentrated, not distributed}

We first measure the atomic-quality probe $q(c_k^{(s)})$ for every ECM
in the library. On T6 (Table~\ref{tab:atomic_t6}) the atomic-quality
matrix is highly concentrated: a single cell, the seed=2024 \textsc{reach}
ECM, achieves an atomic success rate of $88.0\%$, while every other ECM
in the library is at or below $32.0\%$; three of the $16$ cells
sit at exactly $0\%$ and the matrix median is $5.0\%$.

\begin{table}[!ht]
\centering
\resizebox{\linewidth}{!}{%
\begin{tabular}{@{}l cccc@{}}
\toprule
Phase & seed=42 & seed=7 & seed=123 & seed=2024 \\
\midrule
\textsc{reach} & 5.0 [1.0, 10.0]  & 3.0 [0.0, 7.0]   & 27.0 [18.0, 36.0] & \textbf{88.0 [81.0, 94.0]} \\
\textsc{grasp} & 2.0 [0.0, 5.0]   & 14.0 [8.0, 21.0] & 0.0 [0.0, 0.0]    & 1.0 [0.0, 3.0] \\
\textsc{lift}  & 6.0 [2.0, 11.0]  & 0.0 [0.0, 0.0]   & 0.0 [0.0, 0.0]    & 5.0 [1.0, 10.0] \\
\textsc{place} & 32.0 [23.0, 41.0] & 27.0 [18.0, 36.0] & 2.0 [0.0, 5.0]   & 22.0 [14.0, 30.0] \\
\bottomrule
\end{tabular}%
}
\caption{T6 atomic-quality probe $q(c)$ over 100 episodes.}
\label{tab:atomic_t6}
\end{table}

\begin{figure}[!ht]
\centering
\includegraphics[width=0.6\linewidth]{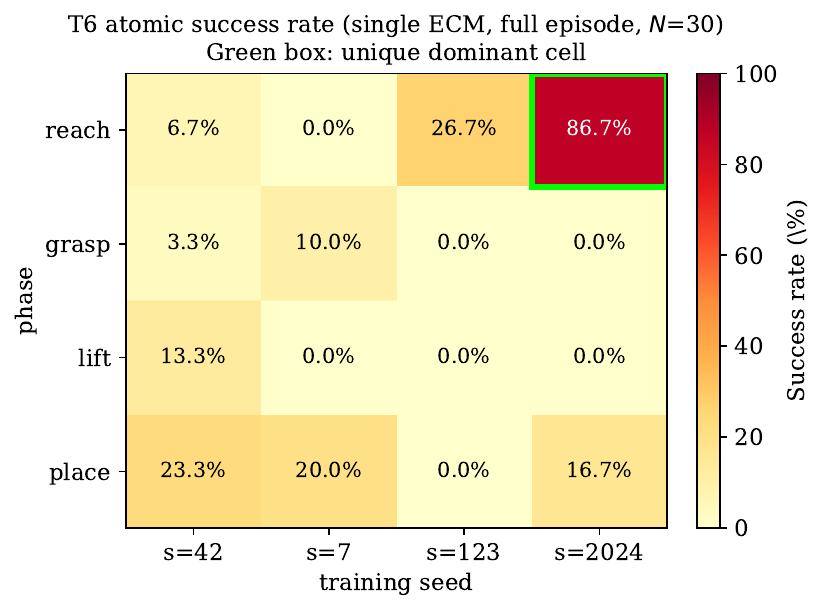}
\caption{T6 atomic success rate per (seed, phase) ECM.}
\label{fig:atomic_t6}
\end{figure}

We refer to the unique highest-$q(\cdot)$ cell as the \emph{dominant
ECM} for that task. With $S{=}4$ we observe a $56$pp gap between
the dominant ($88.0\%$) and the second-best ECM ($32.0\%$) on T6
(visualized in \cref{fig:atomic_t6}). The dominant cell's bootstrap
$95\%$ CI, $[81.0, 94.0]$, is fully disjoint from every other
cell's; the next-best \textsc{reach} cell (seed $123$) has CI
$[18.0, 36.0]$ and the next-best cell overall (\textsc{place},
seed $42$) has $[23.0, 41.0]$ (per-cell CIs in \cref{tab:atomic_t6}).

\subsubsection{Composition success is driven by dominant-ECM inclusion}

Given the atomic concentration, we next ask whether composition outcome
tracks the inclusion of the dominant ECM. For each (primary, alternative)
seed pair on T6, we partition the 16 swap-subsets $\sigma$ into two
groups: those that include the \textsc{reach} phase
(swapping the \textsc{reach} ECM in or out) and those that do not.
Table~\ref{tab:subset_t6} reports the mean composition success rate per
group.

\begin{table}[!ht]
\centering
\resizebox{\linewidth}{!}{%
\begin{tabular}{@{}l rr r@{}}
\toprule
(primary, alt) seed pair
  & \textsc{reach}~$\in \sigma$ [CI]
  & \textsc{reach}~$\notin \sigma$ [CI]
  & $\Delta$ (pp) [CI] \\
\midrule
(42, 2024)    & 70.5\% [67.4, 73.6] & 18.1\% [15.5, 20.9] & \textbf{$+52.4$} [$+48.1, +56.5$] \\
(123, 2024)   & 76.6\% [73.6, 79.6] & 31.1\% [28.0, 34.4] & \textbf{$+45.5$} [$+41.1, +49.8$] \\
(7, 123)      & 34.0\% [30.6, 37.2] & 15.1\% [12.8, 17.6] & $+18.9$ [$+14.8, +22.9$] \\
(42, 7)       & 21.6\% [18.8, 24.5] & 24.2\% [21.2, 27.3] & $\phantom{+}{-2.6}$ [$\phantom{+}{-6.8}, +1.4$] \\
\bottomrule
\end{tabular}%
}
\caption{T6 subset-swap success rate, partitioned by \textsc{reach} membership.}
\label{tab:subset_t6}
\end{table}

The signal is striking: when one of the seeds in the pair has a
high-quality \textsc{reach} ECM (rows 1--3), the composition success
rate moves by +19 to +52pp depending on whether that ECM is in the
swap-set. When neither seed has a high-quality \textsc{reach} ECM
(row 4), the swap-set choice is essentially irrelevant.

\subsubsection{Direction of the effect}

The single-phase paired matrix (\cref{tab:paired_reach_t6}) covers
the $4 \times 4 = 16$ (primary, alternative) seed combinations on the
\textsc{reach} phase, with the diagonal as the within-seed baseline.
The sign-flip predicted by the mechanism is clear: when the primary
\emph{lacks} the dominant ECM (rows $42, 7, 123$), swapping in the
dominant \textsc{reach} (column~$2024$) raises composition success by
$+46$ to $+55$pp; when the primary \emph{has} it (row~$2024$),
swapping out for any of the three alternatives lowers success by
$-31$ to $-59$pp. Column-means span $11.0\%$ to $72.5\%$ and are
governed almost entirely by the atomic quality of the swapped-in
ECM, not by which version was originally present.

\begin{table}[!ht]
\centering
\small
\begin{tabular*}{\linewidth}{@{\extracolsep{\fill}}l rrrr | r}
\toprule
& swap=42 & swap=7 & swap=123 & swap=$\mathbf{2024}$ & diag.\\
\midrule
primary=42        & 16.0 & 13.0 & 28.0 & \textbf{68.0} & 16.0 \\
primary=7         & 29.0 & 19.0 & 48.0 & \textbf{74.0} & 19.0 \\
primary=123       &  8.0 &  0.0 & 31.0 & \textbf{77.0} & 31.0 \\
primary=$\mathbf{2024}$ & 17.0 & 12.0 & 40.0 & 71.0 & \textbf{71.0} \\
\midrule
col.\ mean        & 17.5 & 11.0 & 36.8 & \textbf{72.5} & --- \\
$95\%$ CI         & [\,14.0, 21.2] & [\,8.0, 14.2] & [\,32.2, 41.5] & \textbf{[\,68.2, 76.8]} & --- \\
\bottomrule
\end{tabular*}
\caption{T6 \textsc{reach}-phase paired swap matrix.}
\label{tab:paired_reach_t6}
\end{table}

The pattern is robust across three independent executions of the
same matrix: the original $N{=}30$ GPU-workstation run and an
$N{=}30$ state-logging re-run (used in \cref{sec:disc} for the
mechanism breakdown) give swap=$2024$ column means of $64.2\%$ and
$73.3\%$, against $72.5\%$ in the $N{=}100$ evaluation reported
here; the dominant-column/sibling-column ordering is identical in
all three. The cell-level matrices of all executions are included
in the released artifact bundle.

\subsubsection{Negative controls and atomic predictivity}

A structural alternative (``any swap hurts because composed neural
controllers are inherently fragile'') is ruled out by within-task
negative controls. On T6, the three phases other than \textsc{reach}
lack a high-quality ECM in any seed
(\cref{tab:atomic_t6}: max atomic success rate $\le 32.0\%$, with
\textsc{grasp} and \textsc{lift} at or below $14.0\%$). Swapping ECMs in these phases produces \emph{no} comparable
column-mean shift on the corresponding $4{\times}4$ paired matrix:
on \textsc{grasp}, \textsc{lift}, and \textsc{place}, swap-column
means cluster within $7.8$, $13.0$, and $15.7$pp respectively,
a factor of $3.9$--$7.9$ tighter than the $61.5$pp spread on
\textsc{reach}. The dominant-skill effect therefore requires a true
high-quality ECM, not just a swap event. Numerically, the
\textsc{reach} swap-column means rank $11.0\%$, $17.5\%$, $36.8\%$,
$72.5\%$ in lockstep with the atomic-quality probes
$3.0\%$, $5.0\%$, $27.0\%$, $88.0\%$ of the corresponding seeds: the
atomic probe of the swapped-in ECM monotonically predicts the column
mean of the post-swap composition. The full $4 \times 4$ paired
swap matrices for the three negative-control phases are reported in
\cref{tab:negctrl_all} (\cref{sec:app:negctrl}). One further
signature is visible across all three matrices: the primary=$2024$
rows run $55$--$83\%$ regardless of which phase is swapped, because
the dominant \textsc{reach} ECM remains in those compositions
throughout.

\subsubsection{A controlled weight-space test}

\begin{figure}[!ht]
\centering
\includegraphics[width=\linewidth]{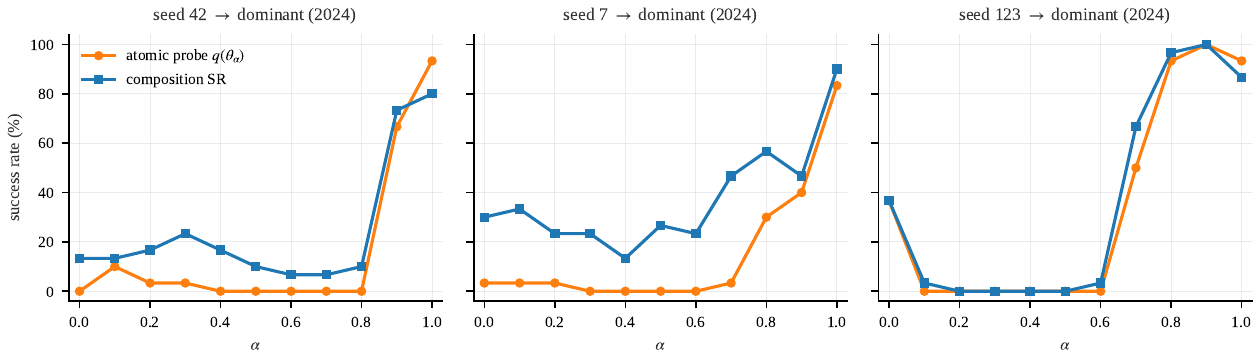}
\caption{Weight-space interpolation between sibling and dominant \textsc{reach} ECMs.}
\label{fig:interp}
\end{figure}

The evidence above is associational: the four candidate ECMs differ
in training seed, so atomic quality co-varies with seed identity.
To manipulate atomic quality directly we interpolate in weight
space: for each sibling seed $s \in \{42, 7, 123\}$ we form the
blended \textsc{reach} controller
$\theta_\alpha = (1{-}\alpha)\,\theta_s + \alpha\,\theta_{2024}$
for $\alpha \in \{0, 0.1, \ldots, 1.0\}$, measure its atomic probe
$q(\theta_\alpha)$, and measure the success rate of the composition
that uses $\theta_\alpha$ at \textsc{reach} with all other phases
fixed to seed $s$ ($N{=}30$ paired episodes per point).
\Cref{fig:interp} shows the result. The paths are far from
monotone: linear interpolation between independently trained SAC
solutions crosses a region where the blended controller fails
outright (atomic SR near $0\%$ for mid-range $\alpha$), a
manifestation of the loss barrier between independent solutions.
Composition success collapses and recovers at the \emph{same}
$\alpha$ values: across all $33$ interpolation points the pooled
Pearson correlation between $q(\theta_\alpha)$ and composition
success is $r{=}0.94$ (per-path $r = 0.98$, $0.92$, $0.99$; the
pooled value is reported descriptively, since points along one
path are serially dependent). The blended mid-path
controllers were never produced by any training run, so the atomic
probe predicts composition outcome for weight configurations
outside the trained candidate set, and the prediction follows the
probe rather than seed identity. This is the controlled
weight-space perturbation that the associational reading calls
for: moving a single phase's weights continuously moves the
composition's deployment quality, point-by-point, in the direction
the atomic probe indicates.

\subsection{Replication on a Second Task: Door}
\label{sec:exp:door}

The dominant-skill effect of \cref{sec:exp:mechanism} rests on a
single task. To test whether it is specific to T6 or a property of
phase-decomposed skill execution more generally, we added a seventh
task, \texttt{T7\_Door} (a contact-rich door-opening task on the same
Panda arm), trained $S{=}4$ phase ECMs per phase under the same seeds,
and repeated the atomic-probe and paired-swap analysis at $N{=}30$.
T7 reproduces the effect, and one of its dominant cells fails to
reproduce it in a way that sharpens the underlying mechanism.

\subsubsection{Atomic quality is again concentrated}
The T7 atomic matrix (\cref{tab:atomic_door}) concentrates quality in
two cells. On \textsc{grasp}, the seed=42 ECM reaches $73.3\%$ atomic
success ($95\%$ CI $[56.7, 90.0]$) while the three siblings stay at or
below $3.3\%$ ($[0.0, 10.0]$): a $70$pp gap with disjoint intervals,
mirroring the concentration seen on T6. On \textsc{place} the seed=123
ECM reaches $80.0\%$ ($[63.3, 93.3]$) against siblings at or below
$20\%$. As on T6, \textsc{reach} saturates (three of four seeds at or
above $96.7\%$), so the dominance ranking is defined on the interior
phases.

\begin{table}[!ht]
\centering
\small
\begin{tabular*}{\linewidth}{@{\extracolsep{\fill}}l rrrr}
\toprule
Phase & seed=42 & seed=7 & seed=123 & seed=2024 \\
\midrule
\textsc{reach} & 100.0\% & 0.0\%  & 96.7\% & 96.7\% \\
\textsc{grasp} & \textbf{73.3\%} & 0.0\% & 0.0\% & 3.3\% \\
\textsc{lift}  & 6.7\%  & 0.0\%  & 36.7\% & 0.0\% \\
\textsc{place} & 20.0\% & 0.0\%  & \textbf{80.0\%} & 0.0\% \\
\bottomrule
\end{tabular*}
\caption{\texttt{T7\_Door} atomic-quality probe $q(c)$ over 30 episodes.}
\label{tab:atomic_door}
\end{table}

\subsubsection{A dominant upstream ECM drives composition}
The \textsc{grasp} paired-swap matrix (\cref{tab:paired_door})
exhibits the conditional sign structure the mechanism predicts, here
carried by the single \textsc{grasp}-bottlenecked primary (seed=7);
on T6 the same logic appeared as a CI-disjoint column-level effect
because \textsc{reach} was the universal bottleneck across primaries,
whereas on T7 only one primary isolates \textsc{grasp}. The
diagnostic row is the
primary seed whose own \textsc{grasp} ECM fails (seed=7, atomic
$0\%$): swapping in the dominant seed=42 \textsc{grasp} ECM raises
composition success from $0\%$ to $83.3\%$ ($[70.0, 96.7]$), while
swapping in any of the three low-quality \textsc{grasp} ECMs leaves it
at $0\%$. On this primary, composition outcome is determined by the
inclusion of the one high-atomic-quality \textsc{grasp} ECM, not by
the swap event. The other three primaries do not isolate
\textsc{grasp}. Seeds 42 and 123 already carry strong \textsc{reach}
ECMs and saturate ($\ge 90\%$) whichever \textsc{grasp} ECM is swapped
in, because \textsc{grasp} is not their bottleneck. Seed 2024 is
non-saturated (its own composition reaches $36.7\%$) but its row does
not track \textsc{grasp} atomic quality: its four swaps span
$36.7$--$53.3\%$ with mutually overlapping bootstrap CIs (the
row-maximal swap-in of seed=123 at $53.3\%$ $[36.7, 70.0]$ versus the
dominant seed=42 at $43.3\%$ $[26.7, 63.3]$), so the dominant
\textsc{grasp} ECM not producing this row's maximum reflects $N{=}30$
sampling noise, and the primary is uninformative about \textsc{grasp}
dominance rather than contradicting it. The positive replication therefore rests on the one primary where
\textsc{grasp} is the binding constraint (seed=7); we treat it as a
single-task replication of the effect, not as a column-level
re-demonstration.

\begin{table}[!ht]
\centering
\small
\begin{tabular*}{\linewidth}{@{\extracolsep{\fill}}l rrrr | r}
\toprule
& swap=$\mathbf{42}$ & swap=7 & swap=123 & swap=2024 & diag.\\
\midrule
primary=$\mathbf{42}$   & 100.0 & 100.0 & 100.0 & 100.0 & 100.0 \\
primary=7               & \textbf{83.3} &  0.0 &  0.0 &  0.0 &  0.0 \\
primary=123             & 100.0 & 96.7  & 100.0 & 90.0  & 100.0 \\
primary=2024            & 43.3  & 36.7  & 53.3  & 36.7  & 36.7 \\
\bottomrule
\end{tabular*}
\caption{\texttt{T7\_Door} \textsc{grasp}-phase paired swap matrix.}
\label{tab:paired_door}
\end{table}

\subsubsection{The dominant phase must be upstream}
T7 supplies a within-task negative case that T6 could not.
\textsc{place} has its own dominant ECM (seed=123, $80.0\%$ atomic),
but it sits at the end of the phase sequence. Swapping this
high-quality \textsc{place} ECM into the failed seed=7 composition does
not rescue it: composition success stays at $0.0\%$, against $83.3\%$
for the upstream \textsc{grasp} swap. A strong terminal-phase ECM
cannot recover an episode that has already failed upstream, because
the composition never reaches a state in which the \textsc{place}
controller acts. The dominant-skill effect therefore requires the
high-quality ECM to lie early enough in the phase sequence to govern
the composition outcome: \textsc{reach} on T6 and \textsc{grasp} on T7
satisfy this, a dominant \textsc{place} ECM does not. This refines the
criterion. The atomic-quality probe predicts composition reliability
for an updated module to the extent that the module's phase lies on
the critical path to task completion.

\subsubsection{Scope of the replication}
T7 replicates the dominant-skill effect and suggests a critical-path
condition on it; it is not a second instance of the full
selector study. We evaluate it at $N{=}30$ rather than $N{=}100$, and
we do not repeat the weight-space interpolation or the Hybrid Selector
cost benchmark on it, which remain T6 results. The evaluation-resolution
caution that applies to selector gold-label verdicts near the
acceptance tolerance (\cref{tab:resolution}) does not weaken the T7
conclusions: they rest on a $70$pp disjoint-CI atomic gap and a
$0\%{\to}83.3\%$ paired jump, both far outside any tolerance band,
where $N{=}30$ is adequate. What T7 establishes is
that the effect underlying the test-selection criterion (one
high-atomic-quality module on the critical path governs composition
success) is not an artifact of a single task. The full T7 atomic,
\textsc{grasp}, and \textsc{place} matrices are in the released
artifact bundle.

\subsection{Boundary: When Atomic Quality Saturates}
\label{sec:exp:boundary}

The dominant-skill effect described in \cref{sec:exp:mechanism}
requires \emph{variation} in atomic skill quality across versions:
some ECMs must be markedly stronger than others. We characterize the
boundary of this effect using a task on which all atomic ECMs are
equally strong.

On T1\_Pick, a single-arm pick of one object from a fixed table
position, every (seed, phase) ECM in our library achieves
$q(c) = 100\%$ atomic success rate (\cref{tab:atomic_t1}). All
four seeds saturate from the first training iteration and hold
$100\%$ across $15$ iterations of the standard schedule; on the
$N{=}30$ atomic-probe evaluation, every one of the
$4 \text{ seeds} \times 4 \text{ phases} = 16$ cells reports
$100\%$ success ($30/30$ episodes per cell, $480$ in total;
\cref{tab:atomic_t1}, \cref{sec:app:negctrl}). The candidate set is
uniformly saturated; there is no dominant cell.

By the mechanism of \cref{sec:exp:mechanism}, when every candidate ECM
contributes equally well, every composition is equally successful and
swapping any phase's ECM cannot shift outcomes. T1 is therefore a
\emph{boundary case} consistent with our mechanism: composition
\emph{is} robust to swap, precisely because there is no atomic-quality
variation for the dominant-skill effect to act on.

\subsection{Why Behavioral Distance Fails to Predict Dominance}
\label{sec:exp:bdist}

A natural alternative hypothesis is that the dominant ECM should be
behaviorally \emph{atypical}: with action distribution distinct from
the others, the dominant ECM might be detectable by a cheap,
model-free distance metric. We test this hypothesis directly. For
each phase we compute the per-ECM mean off-diagonal action $L^2$
distance to its three siblings, evaluated over the paired episode
pool, and ask whether the dominant cell ranks first. Within T6
\textsc{reach}, the dominant ECM (seed=$2024$, atomic SR $88.0\%$)
is rank $3$ of $4$ by mean pairwise distance ($3.349$); the
\emph{most}-distant ECM is the near-zero-SR seed=$7$ at $3.543$;
and the within-phase Spearman correlation between mean pairwise
distance and atomic SR is negative ($\rho = -0.4$; descriptive,
$n{=}4$). Pooled across
all $16$ T6 (seed, phase) cells the correlation is moderate
($\rho = 0.70$, bootstrap $95\%$ CI $[0.27, 0.90]$), yet the
ranking it induces still misses the dominant module: the dominant
cell sits at rank $7$ of $16$ by mean pairwise distance, so a
top-$1$ distance rule fails both within its phase and across the
matrix. On T3 and T4
atomic SR is degenerate at $0\%$ on nearly every cell, so the
within-task ranking carries no information and the refutation is
defined only on T6. The combined evidence: behavioral distance is
a poor detector of the dominant cell: it places it near the
middle of its phase, not at the top. \Cref{tab:bdist} reports the per-(task, phase) off-diagonal
summary; the full $12$-panel pairwise distance grid is visualized
in \cref{fig:bdist} (\cref{sec:app:bdist}).

\begin{table}[!ht]
\centering
\small
\begin{tabular*}{\linewidth}{@{\extracolsep{\fill}}l rrrr}
\toprule
Task & \textsc{reach} & \textsc{grasp} & \textsc{lift} & \textsc{place} \\
\midrule
T6 & 3.40 & 3.00 & 3.07 & 3.51 \\
T3 & 2.71 & 2.66 & 2.59 & 1.90 \\
T4 & 1.51 & 1.06 & 1.81 & 0.94 \\
\bottomrule
\end{tabular*}
\caption{Mean off-diagonal action $L^2$ distance per (task, phase).}
\label{tab:bdist}
\end{table}

\subsection{Hybrid Selector Benchmark: Cost-vs-Safety Trade-off}
\label{sec:exp:selector}

We now benchmark the selectors defined in \cref{sec:method:selectors}
against the gold label. The selector definitions, the Hybrid Selector
pseudocode (\cref{alg:hybrid}), and per-decision cost analysis all
appear in \cref{sec:method:selectors}; this section reports the
empirical cost--quality trade-off.

\subsubsection{Results on T6 (success-rate oracle)}
\Cref{tab:algo_t6} reports gold-label match at $\tau{=}5$pp on T6
across 48 update events ($4{\times}3$ ordered seed pairs $\times$ 4
phases), with every probe evaluated at $N{=}100$ paired episodes.
The zero-cost AtomicOnly selector matches the gold label on
$36/48 = 75.0\%$ of events, the same count as FullReval
($36/48$): the
$2 \times 2$ disagreement table is symmetric (both correct $32$,
AtomicOnly-only $b{=}4$, FullReval-only $c{=}4$;
McNemar exact $p{=}1.0$). With only $8$ discordant events the test
has power against large differences only, so we read this as no
detectable difference rather than demonstrated equivalence. Hybrid$(m{=}10)$ triggers FullReval on
$24$ of $48$ events ($50\%$ cost) and reaches $39/48 = 81.25\%$,
the best match of all seven selectors; it never loses to FullReval
on any event (wins $3$, loses $0$; McNemar $p{=}0.25$), and of the
$16$ (phase $\times$ new-seed) update clusters none favours
FullReval over it ($3$ favour the Hybrid, $13$ tie; cluster
permutation $p{=}0.25$, $B{=}5000$). Against a Random selector
that spends the identical FullReval budget on uniformly chosen
events (applying the AtomicOnly rule on the rest; $4000$ MC
iterations), the Hybrid's match rate is significant: Random reaches
$75.1\%$ on average ($95\%$ MC interval $[68.8, 81.2]$) and
attains the Hybrid's $81.25\%$ or better with probability
$0.039$.

\subsubsection{Evaluation resolution changes the conclusion}
The original version of this experiment, evaluated at $N{=}30$
episodes per probe, told a different story: AtomicOnly $64.6\%$,
Hybrid($m{=}10$) $75.0\%$, FullReval $87.5\%$, with the
AtomicOnly-vs-FullReval gap significant at $p{=}0.013$. A
fixed-prefix $N{=}30$ subsample of the present data reproduces
that pattern ($64.6\%$ / $75.0\%$ / $85.4\%$), so the reversal is
driven by evaluation resolution, not by platform or protocol
(\cref{tab:resolution}). The mechanism is the interaction between
gold-label granularity and the tolerance band: at $N{=}30$ the
success-rate grid is $3.33$pp coarse and few measured drops land
inside the $\tau$-acceptance band, so FullReval
rarely disagrees with the zero-tolerance gold label; at $N{=}100$
the grid is $1$pp fine, $12$ of $48$ events land in the band
(eleven strict drops and one exact tie), and
FullReval mismatches the gold label on every one of them
($25.0\%$ unsafe, \cref{sec:exp:safety}). Full revalidation's
apparent advantage at coarse resolution is an artifact of
measurement granularity, and test-selection studies that benchmark
against an empirical revalidation oracle should report the
evaluation resolution alongside the verdicts.

\begin{table}[!ht]
\centering
\small
\begin{tabular*}{\linewidth}{@{\extracolsep{\fill}}l cc}
\toprule
Selector & match\,\% at $N{=}30$ prefix & match\,\% at $N{=}100$ \\
\midrule
AtomicOnly       & 64.6 & 75.0 \\
Hybrid($m{=}10$) & 75.0 & \textbf{81.2} \\
FullReval        & \textbf{85.4} & 75.0 \\
\bottomrule
\end{tabular*}
\caption{Resolution sensitivity of the selector benchmark ($N{=}30$ versus $N{=}100$ gold labels).}
\label{tab:resolution}
\end{table}

\subsubsection{Gold-label robustness}
The gold label above is computed from the same $N{=}100$ episode
pool the selectors observe, which makes FullReval's band
mismatches definitional rather than sampled. Two checks show the
conclusions are not artifacts of this convention. First, a
split-half design separates the pools: selectors observe episodes
$1$--$50$ and the gold label is computed from episodes
$51$--$100$, and vice versa. In both directions FullReval's
advantage over the zero-cost probe is at most a single event
(match $64.6\%$ vs $70.8\%$ in one direction, $75.0\%$ vs
$72.9\%$ in the other), and Hybrid($m{=}10$) matches or improves
on every alternative's unsafe rate and fault-detection sensitivity
($20.8\%$/$60.0\%$ and $16.7\%$/$61.9\%$, the latter pair tying
FullReval); the Hybrid's own match rates ($68.8\%$ and $77.1\%$)
sit within half-to-half variability and trail AtomicOnly in one
direction. Second, sweeping
the tolerance: at $\tau \in \{2, 5, 10\}$pp the Hybrid holds the
highest full-pool match ($85.4$/$81.2$/$79.2\%$, sharing the
maximum with FullReval at $\tau{=}2$) while FullReval
decays from $85.4\%$ to $70.8\%$ as the band widens and
AtomicOnly stays near $75\%$ throughout; the $\tau{=}0$ column is
degenerate by construction, since FullReval then differs from the
gold label only on exact ties. The qualitative conclusions (the
zero-cost probe loses nothing relative to full revalidation, and
margin-gating buys safety per unit of budget) hold under every
convention tested; the superlative ``best match'' is specific to
the full-pool gold label at $\tau{=}5$.

\subsubsection{Code-level RTS baseline}
For completeness we instantiate an Ekstazi-style file-dependency
selector~\citep{gligoric2015ekstazi} on the same $48$ events. At
source granularity it selects nothing: the training code is
byte-identical across ECM versions, the dependency diff is empty,
and the selector reduces to Naive ($50.0\%$ match, $50.0\%$
unsafe). At weight-file granularity it selects everything: the
checkpoint hash changes for every update event by construction,
every dependent composition is re-tested, and the selector reduces
to FullReval at cost $1.0$. Either granularity sits at an endpoint
of the cost axis; neither can route budget by expected behavioral
impact, which is the capability the atomic margin provides.

\begin{table}[!ht]
\centering
\small
\begin{tabular*}{\linewidth}{@{\extracolsep{\fill}}l rr r}
\toprule
Selector & Gold-label match [$95\%$ CI] & Cost \\
\midrule
Naive (accept all)                      & 50.0\% [\,35.4, 64.6] & 0\%   \\
Freeze (reject all; BLADE/SymSkill)     & 50.0\% [\,35.4, 64.6] & 0\%   \\
AtomicOnly                              & 75.0\% [\,62.5, 87.5] & 0\%   \\
FullReval                               & 75.0\% [\,62.5, 85.4] & 100\% \\
\textbf{Hybrid($m{=}10$)}               & \textbf{81.2\%} [\,68.8, 91.7] & 50.0\%  \\
Hybrid($m{=}20$)                        & 77.1\% [\,64.6, 87.5] & 66.7\%  \\
Hybrid($m{=}30$)                        & 77.1\% [\,64.6, 87.5] & 83.3\%  \\
\bottomrule
\end{tabular*}
\caption{T6 gold-label match rate at $N{=}100$ evaluation.}
\label{tab:algo_t6}
\end{table}

\subsubsection{Cross-task pattern (mixed-oracle caveat)}

Extending the benchmark to T3 and T4 introduces a methodological
subtlety: on these two tasks all atomic policies achieve $0\%$
task-success rate (\cref{sec:lim}), so the oracle there is defined
on \emph{reward} instead (\cref{sec:exp:tasks}). Because a
reward-oracle match and an SR-oracle match are not directly
comparable, we
treat the $96$ reward-oracled events as an exploratory extension
and report the full per-task and aggregate tables in
\cref{sec:app:xtask}. Two observations carry over: AtomicOnly
(cost~$0$) is within $3$pp of FullReval (cost~$100\%$) on the
aggregate, and on T3 and T4 specifically AtomicOnly \emph{beats}
FullReval, reflecting that when composition signals are noisy or
collapsed a per-skill probe can be cleaner than a per-composition
probe. On the aggregate, however, the Hybrid's separation from
FullReval is negligible (\cref{sec:app:xtask}); the T6 results
above are the methodologically conservative evidence for the
Hybrid Selector. We report bootstrap $95\%$ CIs throughout,
following~\citet{agarwal2021precipice}'s recommendations for
sparse-trial RL benchmarks.

\subsection{Safety: False-Negative and False-Positive Rates}
\label{sec:exp:safety}

The oracle-match rate reported above conflates two error modes that
the regression-testing literature
\citep{rothermel1996analyzing,yoo2012regression} keeps separate. A
\emph{false-negative} (unsafe) disagreement is a missed regression:
full revalidation would have flagged a faulty composition; the
criterion would not. A \emph{false-positive} (conservative)
disagreement is a spurious failure: the criterion flags a
composition that full revalidation would have passed. These are not
symmetric under deployment use: false negatives release
silently-broken compositions to production, while false positives
only inflate test cost. \Cref{tab:safety_split} reports the split for
each strategy on T6, where the success-rate oracle is well-defined;
the mixed-oracle cross-task aggregate is reported in
\cref{tab:safety_aggregate} (\cref{sec:app:xtask}).

\begin{table}[!ht]
\centering
\caption{Safety decomposition of the T6 gold-label match rate.}
\label{tab:safety_split}
\begin{tabular*}{\linewidth}{@{\extracolsep{\fill}}l rrrrr}
\toprule
                    & Match\,\% & Unsafe\,\% & Cons\,\% & Sens\,\% & Cost \\
\midrule
Naive                       & 50.00 & 50.00 &  0.00 &   0.0 & 0.000 \\
AtomicOnly                  & 75.00 & 22.92 &  2.08 &  54.2 & 0.000 \\
Hybrid($m{=}10$)            & 81.25 & 18.75 &  0.00 &  62.5 & 0.500 \\
Hybrid($m{=}20$)            & 77.08 & 22.92 &  0.00 &  54.2 & 0.667 \\
Hybrid($m{=}30$)            & 77.08 & 22.92 &  0.00 &  54.2 & 0.833 \\
Random@cost$=0.500$         & 75.1  & 23.9  &  1.0  &  52.2 & 0.500 \\
FullReval                   & 75.00 & 25.00 &  0.00 &  50.0 & 1.000 \\
\bottomrule
\end{tabular*}
\end{table}

\begin{figure}[!ht]
\centering
\includegraphics[width=\linewidth]{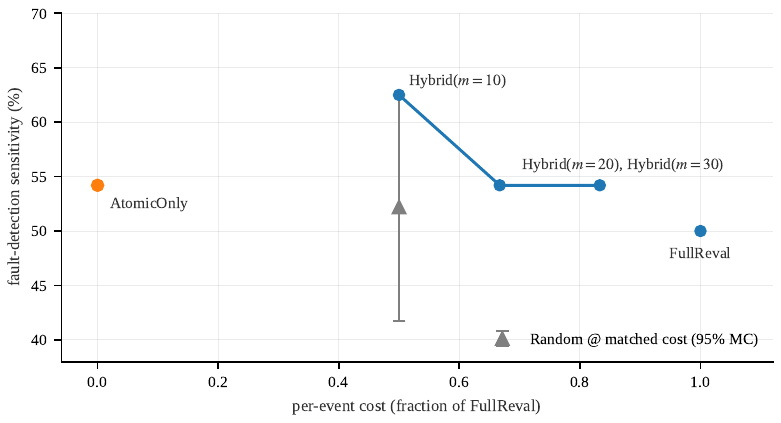}
\caption{Fault-detection sensitivity versus per-event cost on T6.}
\label{fig:sens_vs_cost}
\end{figure}

Three findings stand out. First, Hybrid($m{=}10$) attains both the
lowest unsafe rate of the budget-spending selectors ($18.75\%$,
against $22.92\%$ for AtomicOnly and $25.0\%$ for FullReval) and
the highest fault-detection sensitivity ($62.5\%$), at half of
FullReval's cost; Freeze trivially catches every regression by
rejecting everything and is excluded from this comparison. Second,
the $25.0\%$ unsafe rate on FullReval itself is informative: it
counts the twelve events whose measured drop falls inside the
$\tau{=}5$pp acceptance band (including one exact tie on the
$N{=}100$ pool), and it therefore bounds from below the unsafe
rate of any $\tau$-tolerant criterion operating on the same
observable. The underlying verdicts
are finite-sample Monte-Carlo estimates: re-executing the same
composition on a fresh episode pool can flip a verdict near the
acceptance boundary, the compositional-ML analogue of a flaky
test~\citep{luo2014flaky}, and the tolerance band is the mechanism
that absorbs such verdict instability at a measured safety cost;
the resolution analysis of \cref{sec:exp:selector} shows that this
band is exactly what coarse evaluation hides.
The single exact tie (\textsc{lift}, $2024 \to 123$) does not
drive any conclusion: relabeling it as a pass lifts AtomicOnly and
FullReval to $77.08\%$ and Hybrid($m{=}10$) to $83.33\%$ with the
ordering unchanged.
Third, comparison
against a Random selector at exactly matched cost ($0.50$;
Monte-Carlo estimate, $4000$ iterations) shows the margin is doing
real routing work: Random reaches $75.1\%$ match against the
Hybrid's $81.25\%$ ($P(\mathrm{Random} \geq \mathrm{Hybrid}) =
0.039$) and $52.2\%$ sensitivity against the Hybrid's $62.5\%$
($P = 0.066$). The match advantage is significant, the sensitivity
advantage directional; both are consistent with the margin-gated
criterion routing scarce FullReval budget toward events where it
reduces missed regressions, the regression-testing property the
match rate alone cannot
expose. We do not report
APFD~\citep{elbaum2000prioritizing,rothermel2001tcp} as a primary
metric here: APFD presumes an ordered test suite, while our
selectors emit independent per-composition decisions; the
fault-detection sensitivity reported above is the natural analogue.
The corresponding decomposition on the mixed-oracle cross-task
aggregate is reported in \cref{sec:app:xtask}.

\section{Discussion}
\label{sec:disc}

\subsection{The Dominant-Skill Effect as a Per-Skill Regression-on-Update Phenomenon}
We interpret the dominant-skill effect as the per-skill instance of
the regression-on-update phenomenon studied in the
backward-compatibility literature
\citep{bansal2019backwardcompat,shen2020backwardcompat,yan2021pct}:
even when an updated function is on average no worse, it can flip
individual downstream decisions in ways the old function did not.
The same asymmetry explains the vanishing population mean: swapping
toward and away from the dominant ECM produce gains and losses of
comparable magnitude, so the signal lives in the conditional
structure rather than the marginal mean. A naive methodology that
reports population-mean $\Delta$ under swap will therefore
systematically miss the phenomenon.

\subsection{Alternative Mechanisms Tested Directly}
A natural prior is that a high-quality atomic ECM is robust across
a wider range of hand-off state distributions than its
lower-quality siblings. We tested three refinements of this
robustness-asymmetry story on T6 \textsc{reach}:
\textbf{(a)~hand-off state coverage}: the dominant ECM may visit a
wider region of phase-end states, so any downstream phase finds
itself in-distribution;
\textbf{(b)~action smoothness}: smoother action trajectories reduce
contact discontinuities at phase transitions, an effect related to
T-STAR's terminal-state regularization~\citep{lee2021tstar};
\textbf{(c)~trajectory-length distribution}: a dominant ECM may
finish its phase faster (or slower) on average, leaving the
downstream phase a larger time budget.

We measured~(a) directly. For each of the four T6 \textsc{reach}
ECMs we collected the phase-end state vector ($\dim{=}218$) over
$N{=}120$ episodes (30 episodes $\times$ 4 swap configurations) and
computed pairwise Wasserstein-2 distance under a diagonal-Gaussian
approximation, the L${}^2$ shift to the pooled centroid, and the
sum of per-dimension variances ($\sum \mathrm{diag}\,\mathrm{Cov}$).
The dominant ECM (seed${=}2024$) is \emph{neither shifted further
from the pooled centroid nor wider} than its siblings
(\cref{tab:mechanisms}, rows~(a)).
The mean pairwise W${}^2$ between the dominant and the
three siblings is $28.1$, smaller than the sibling-to-sibling mean
of $34.7$; the dominant distribution is $-23.6\%$ less shifted
from the pooled centroid and $-29.7\%$ narrower than the sibling
mean, with greater per-dimension variance on only $4.6\%$ of the
$218$ dimensions. Mechanism~(a) is therefore \emph{not} supported
in the present data: the dominant ECM's phase-end state distribution
is, if anything, more concentrated and more central.

We then measured (b) and (c) on a follow-up rollout of the same
$16$ configurations with per-step action and per-step state logging
enabled ($120$ episodes per ECM, reach phase truncated at the
framework boundary $T{=}62$ steps). Mechanism~(b) is the per-episode
mean step-to-step action change
$\langle\|\mathbf{a}_t - \mathbf{a}_{t-1}\|_2\rangle_t$;
mechanism~(c) is the per-episode L${}^2$ path length through state
space $\sum_t \|\mathbf{s}_{t+1} - \mathbf{s}_t\|_2$. Lower values
on either metric indicate the proposed mechanism (smoother / more
efficient); \cref{tab:mechanisms} reports all three channels.

\begin{table}[!ht]
\centering
\small
\begin{tabular*}{\linewidth}{@{\extracolsep{\fill}}l rrrr}
\toprule
seed & 42 & 7 & 123 & \textbf{2024 (dominant)}\\
\midrule
atomic SR (\%)
            & 5.0  & 3.0  & 27.0 & \textbf{88.0}\\
\midrule
(a) $\sum \mathrm{diag}\,\mathrm{Cov}$ (width, $\uparrow$ predicted)
            & \textbf{3443} & 3181 &  357 & 1635\\
(a) $\|\mu_s - \bar{\mu}\|_2$ (shift, $\uparrow$ predicted)
            & \textbf{7.10} & 5.81 & 5.55 & 4.70\\
(b) action smoothness ($\downarrow$ predicted)
            & 2.27 & 2.98 & \textbf{0.76} & 1.48\\
(c) traj.\ length ($\downarrow$ predicted)
            & 4567 & 4911 & \textbf{2787} & 3855\\
\bottomrule
\end{tabular*}
\caption{Alternative-mechanism measurements for the four T6 \textsc{reach} ECMs.}
\label{tab:mechanisms}
\end{table}

\noindent On both~(b) and~(c) the dominant ECM is \emph{not} the
extreme: it ranks $2$ of $4$ on each metric, with the lower-quality
seed=$123$ (atomic SR $27.0\%$) holding the smoothest and shortest
spots. Pairwise bootstrap $95\%$ CIs confirm the dominant ECM is
significantly smoother than seeds=$42$ and~$7$ (mean $\Delta {<} 0$,
CI excludes zero) but significantly rougher than seed=$123$
(mean $\Delta {>} 0$, CI excludes zero); the same pattern holds for
trajectory length. The conclusion matches~(a): \emph{no single
channel} of (a) hand-off coverage, (b) action smoothness, or
(c) trajectory length identifies the dominant ECM as an outlier on
its own. A direct joint test (\cref{sec:app:joint}) sharpens the
conclusion: pairing the channels adds nothing under a linear
read-out (joint area under the ROC curve, AUC, $0.68$ vs $0.67$ for the best marginal), and
although a nonlinear read-out can distinguish every ECM from every
other (each policy carries its own behavioral signature, $k$NN AUC
$0.94$--$1.00$), dominance does not align with behavioral
atypicality: the dominant ECM ranks \emph{last} of the four under
both linear and nonlinear atypicality scores. An unsupervised
detector of behavioral unusualness would flag a $0\%$-success
sibling, not the dominant module; what the atomic probe measures
is therefore not behavioral atypicality, and identifying its
structural basis is open future work.

\subsection{Connection to Typed-Composition Literature and Deployment}
Our findings reinforce the typed-composition thread
\citep{liu2024blade,shao2025symskill,mishra2023gsc} that pre/post-
condition structure is the right level at which to reason about
compositional behavior. Where prior work stops at \emph{constructing}
such structure, we add a complementary observation: even with
identical type signatures, two ECM versions can produce dramatically
different composition outcomes, and the difference is captured by an
atomic-quality probe rather than by any structural metric. Concretely
for deployment, every candidate skill update should be probed
atomically \emph{first}; composition probes (far more expensive)
should be invoked only when the atomic margin is insufficient. At
the evaluation resolution of this paper that policy is not merely
cheaper: it yields the best gold-label agreement of every selector
benchmarked, because the atomic margins are large exactly where
the probe is trusted, leaving it far less exposed to the tolerance
band that erodes the composition probe's own verdicts.

\section{Additional Analyses}
\label{sec:additional}

This section collects supporting analyses referenced in the main text: per-cell confidence intervals, the within-task negative-control swap matrices and saturation data, the behavioral-distance grid, the exploratory cross-task reward-oracle extension, and the joint behavioral-channel test.

\subsection{Per-Cell Confidence Intervals}
\label{sec:app:ci}

This section tabulates the per-cell bootstrap $95\%$ confidence
intervals (B${=}5000$, $N{=}100$ paired episodes per cell)
underlying the point estimates of
\cref{tab:atomic_t6,tab:paired_reach_t6}.

\begin{table}[!ht]
\centering
\resizebox{\linewidth}{!}{%
\begin{tabular}{@{}l rrrr@{}}
\toprule
& swap=42 & swap=7 & swap=123 & swap=2024 \\
\midrule
primary=42  & 16.0 [\phantom{0}9.0,23.0] & 13.0 [\phantom{0}7.0,20.0] & 28.0 [19.0,37.0] & \textbf{68.0 [59.0,77.0]} \\
primary=7   & 29.0 [20.0,38.0] & 19.0 [12.0,27.0] & 48.0 [38.0,58.0] & \textbf{74.0 [65.0,82.0]} \\
primary=123 & \phantom{0}8.0 [\phantom{0}3.0,14.0] & \phantom{0}0.0 [\phantom{0}0.0,\phantom{0}0.0] & 31.0 [22.0,40.0] & \textbf{77.0 [68.0,85.0]} \\
primary=2024 & 17.0 [10.0,25.0] & 12.0 [\phantom{0}6.0,19.0] & 40.0 [30.0,50.0] & 71.0 [62.0,80.0] \\
\bottomrule
\end{tabular}%
}
\caption{Per-cell T6 \textsc{reach}-phase paired-swap success rate with $95\%$ CIs.}
\label{tab:paired_reach_t6_ci}
\end{table}

\subsection{Negative-Control Swap Matrices and Saturation Data}
\label{sec:app:negctrl}

\Cref{tab:negctrl_all} reports the full $4 \times 4$ paired swap
matrices for the three T6 phases that lack a high-quality ECM
(\cref{sec:exp:mechanism}); \cref{tab:atomic_t1} reports the
saturated T1 atomic-probe matrix underlying the boundary case
(\cref{sec:exp:boundary}).

\begin{table}[!ht]
\centering
\small
\begin{tabular*}{\linewidth}{@{\extracolsep{\fill}}l rrrr}
\toprule
 & swap=42 & swap=7 & swap=123 & swap=2024 \\
\midrule
\multicolumn{5}{l}{\textsc{grasp} (column spread $7.8$pp)} \\
primary=42        & 16.0 & 30.0 & 24.0 & 24.0 \\
primary=7         & 26.0 & 36.0 & 32.0 & 30.0 \\
primary=123       & 31.0 & 28.0 & 22.0 & 41.0 \\
primary=2024      & 63.0 & 73.0 & 83.0 & 65.0 \\
column mean       & 34.0 & 41.8 & 40.2 & 40.0 \\
\midrule
\multicolumn{5}{l}{\textsc{lift} (column spread $13.0$pp)} \\
primary=42        & 17.0 & 24.0 & 16.0 & 16.0 \\
primary=7         & 17.0 & 37.0 & \phantom{0}8.0 & 22.0 \\
primary=123       & 33.0 & 33.0 & 24.0 & 30.0 \\
primary=2024      & 75.0 & 78.0 & 72.0 & 72.0 \\
column mean       & 35.5 & 43.0 & 30.0 & 35.0 \\
\midrule
\multicolumn{5}{l}{\textsc{place} (column spread $15.7$pp)} \\
primary=42        & 21.0 & 23.0 & 18.0 & 19.0 \\
primary=7         & 23.0 & 28.0 & \phantom{0}7.0 & 27.0 \\
primary=123       & 34.0 & 37.0 & 27.0 & 25.0 \\
primary=2024      & 65.0 & 82.0 & 55.0 & 68.0 \\
column mean       & 35.8 & 42.5 & 26.8 & 34.8 \\
\bottomrule
\end{tabular*}
\caption{T6 negative-control paired swap matrices for the three non-dominant phases.}
\label{tab:negctrl_all}
\end{table}

\begin{table}[!ht]
\centering
\small
\begin{tabular*}{\linewidth}{@{\extracolsep{\fill}}l c c c c}
\toprule
Phase & seed=42 & seed=7 & seed=123 & seed=2024 \\
\midrule
\textsc{reach} & 100\% & 100\% & 100\% & 100\% \\
\textsc{grasp} & 100\% & 100\% & 100\% & 100\% \\
\textsc{lift}  & 100\% & 100\% & 100\% & 100\% \\
\textsc{place} & 100\% & 100\% & 100\% & 100\% \\
\bottomrule
\end{tabular*}
\caption{T1 atomic-quality probe across all four seeds.}
\label{tab:atomic_t1}
\end{table}

\subsection{Behavioral-Distance Grid}
\label{sec:app:bdist}

\begin{figure}[!ht]
\centering
\includegraphics[width=0.85\linewidth]{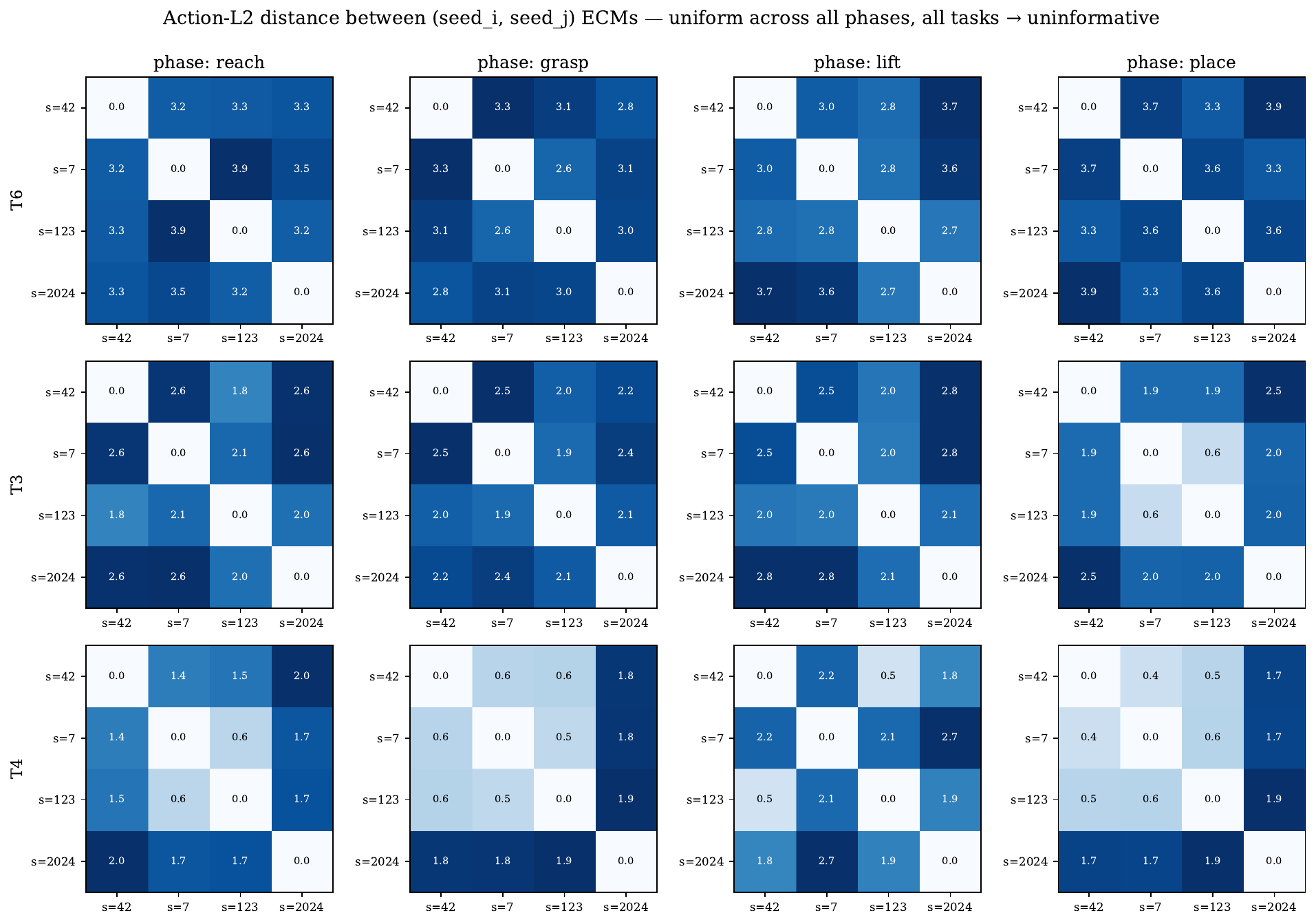}
\caption{Pairwise action $L^2$ distance between ECMs, per (task, phase).}
\label{fig:bdist}
\end{figure}

\Cref{fig:bdist} shows the full pairwise action $L^2$ distance grid
underlying the per-(task, phase) summary of \cref{tab:bdist}. All
twelve panels are visually uniform, and the dominant T6
\textsc{reach} cell sits within the typical range of its siblings.

\subsection{Cross-Task Reward-Oracle Extension}
\label{sec:app:xtask}

\begin{figure}[!ht]
\centering
\includegraphics[width=\linewidth]{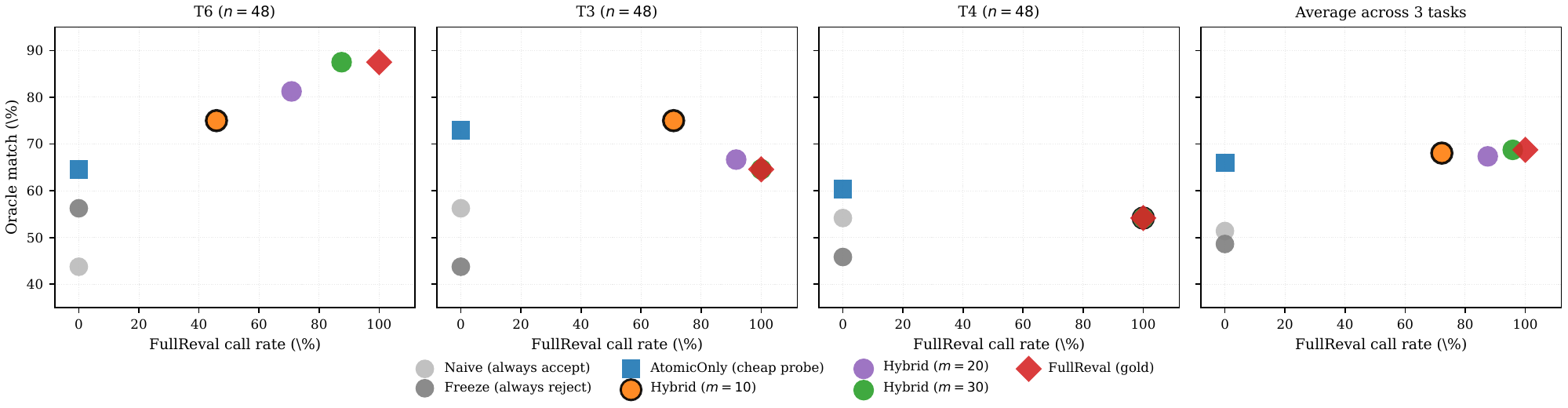}
\caption{Cost--accuracy Pareto frontier for the seven selectors.}
\label{fig:algo_pareto}
\end{figure}

The full cross-task selector benchmark introduced in
\cref{sec:exp:selector} is reported here. On T3 and T4 all atomic
policies achieve $0\%$ task success, so the oracle and all probes
substitute mean episode return for the success rate
(\cref{sec:exp:tasks}); the $144$-event aggregate therefore mixes
two oracles on different scales and is reported as an exploratory
extension of the success-rate-oracled T6 evidence, not as primary
support. The entire extension retains the original $N{=}30$
evaluation, including its T6 column; the main-text T6 results use
$N{=}100$, and \cref{tab:resolution} reports how the two
resolutions relate. \Cref{fig:algo_pareto} plots the corresponding
cost--accuracy frontier per task and on the aggregate.

\begin{table}[!ht]
\centering
\small
\begin{tabular*}{\linewidth}{@{\extracolsep{\fill}}l c c c c}
\toprule
Selector & T6 (SR) & T3 (rew) & T4 (rew) & Avg.\ ($n{=}144$)\\
\midrule
Naive       & 43.8 [\,29, 58] & 56.2 [\,42, 71] & 54.2 [\,40, 69] & 51.4 [\,43, 60] \\
Freeze      & 56.2 [\,42, 69] & 43.8 [\,29, 58] & 45.8 [\,31, 60] & 48.6 [\,40, 57] \\
AtomicOnly  & 64.6 [\,52, 77] & 72.9 [\,60, 85] & 60.4 [\,46, 73] & 66.0 [\,58, 74] \\
FullReval   & 87.5 [\,77, 96] & 64.6 [\,52, 77] & 54.2 [\,40, 69] & 68.8 [\,61, 76] \\
Hybrid($m{=}10$) & 75.0 [\,62, 85] & 75.0 [\,62, 88] & 54.2 [\,40, 69] & 68.1 [\,60, 76] \\
Hybrid($m{=}20$) & 81.2 [\,68, 90] & 66.7 [\,53, 78] & 54.2 [\,40, 67] & 67.4 [\,59, 74] \\
Hybrid($m{=}30$) & 87.5 [\,75, 94] & 64.6 [\,50, 77] & 54.2 [\,40, 67] & 68.8 [\,61, 76] \\
\bottomrule
\end{tabular*}
\caption{Per-task and cross-task oracle match rates.}
\label{tab:algo_xtask}
\end{table}

\begin{table}[!ht]
\centering
\caption{Safety decomposition on the 144-event cross-task aggregate.}
\label{tab:safety_aggregate}
\begin{tabular*}{\linewidth}{@{\extracolsep{\fill}}l rrrrr}
\toprule
                    & Match\,\% & Unsafe\,\% & Cons\,\% & Sens\,\% & Cost \\
\midrule
Naive                       & 51.39 & 48.61 &  0.00 &   0.0 & 0.000 \\
AtomicOnly                  & 65.97 & 31.94 &  2.08 &  34.3 & 0.000 \\
Hybrid($m{=}10$)            & 68.06 & 29.86 &  2.08 &  38.6 & 0.722 \\
Hybrid($m{=}20$)            & 67.36 & 31.94 &  0.69 &  34.3 & 0.875 \\
Hybrid($m{=}30$)            & 68.75 & 31.25 &  0.00 &  35.7 & 0.958 \\
Random@cost$=0.722$         & 67.9  & 31.5  &  0.6  &  35.3 & 0.722 \\
FullReval                   & 68.75 & 31.25 &  0.00 &  35.7 & 1.000 \\
\bottomrule
\end{tabular*}
\end{table}

On this aggregate the Hybrid selectors do not separate from
FullReval (Hybrid($m{=}30$) and FullReval both reach $68.75\%$
match), and the Hybrid's fault-detection-sensitivity advantage over
Random at matched cost is small ($38.6\%$ vs $35.3\%$, inside the
Random $95\%$ MC interval). Both observations follow from the
reward-oracled events, on which every atomic policy scores $0\%$
task success (\cref{sec:exp:tasks}): with no atomic-quality
variation, margin-gating has no success-rate signal to route
FullReval budget by. The T6 split (\cref{tab:safety_split}) is
where the criterion has signal to act on, which is why the paper's
claims are scoped to the success-rate-oracled setting.

\subsection{Joint Behavioral-Channel Test}
\label{sec:app:joint}

\Cref{sec:disc} reports that none of the three behavioral channels
(hand-off coverage, action smoothness, trajectory length) identifies
the dominant ECM on its own. This section tests the natural
follow-up: whether a \emph{joint} statistic over those channels
recovers the dominance signal that the marginals miss.

From the instrumented re-rollout of the $16$ T6 \textsc{reach}
configurations (per-step actions and states; $N{=}120$ episodes per
reach ECM), we compute three per-episode features: mean step-to-step
action change (smoothness), L${}^2$ distance of the phase-end state
to the pooled centroid (coverage centrality), and state-space path
length. Episodes group by the swapped-in reach ECM.
Feature-level dependence between matched episodes (same
initial-state seed, different primary) is negligible: matched-pair
correlations lie in $[-0.32, 0.37]$ and episode-seed intraclass
correlations lie between $-0.06$ and $-0.01$ across the three
features, so all $120$ episodes per ECM enter the analysis, with
cross-validation folds grouped by initial-state seed to rule out
leakage.

For the dominant ECM against its three pooled siblings,
location-based marginal discriminability is weak on every channel
(Mann--Whitney AUC: smoothness $0.67$, centrality $0.64$, path
length $0.61$), and pairing the channels adds nothing under a
linear read-out: the joint smoothness$\times$centrality Fisher-LDA
score reaches AUC $0.68$ against $0.67$ for the best marginal
under identical grouped folds (gain $+0.007$); adding path length
reaches $0.76$. A nonlinear read-out changes the picture in an
instructive way: a $k$-nearest-neighbour score on the same two
features reaches AUC $0.94$, and $0.94$ on smoothness alone,
because the dominant ECM occupies a narrow mid-range smoothness
band between its siblings' modes. The behavioral channels
therefore carry enough signal to distinguish every ECM from every
other (each policy has its own signature), and the precise
negative result is sharper than ``no signal'': dominance does not
align with behavioral \emph{atypicality}. Ranking the four ECMs by
one-vs-rest separability places the dominant ECM \emph{last} under
both read-outs (linear: $0.68$ against siblings at
$0.70$--$0.99$; $k$NN: $0.94$ against siblings at
$0.97$--$1.00$), so an unsupervised detector that flags the
behaviorally most unusual candidate would select a $0\%$-success
sibling and never the dominant module. The analysis script, the
per-episode feature table, and all numbers above are included in
the released artifact.


\section{Limitations and Future Work}
\label{sec:lim}

We organize the limitations under the four validity dimensions
standard in empirical software engineering~\citep{yoo2012regression}:
construct, internal, external, and conclusion validity.

\subsection{Construct Validity}
The atomic-quality probe $q(c)$ measures per-ECM Monte-Carlo success
rate on its phase, while we use it as a \emph{predictor} of
composition reliability under update. The construct gap is real:
$q$ is a per-module quantity, the property we want to test
(post-update composition success) is a downstream interaction. The
gap is bounded empirically by the agreement rates reported in
\cref{tab:safety_split}, but a probe that more directly measures
``downstream composability'' (for example a learned residual that
predicts composition success from ECM behavior) could in principle
narrow it further. A second construct caveat concerns the oracle
itself. The cross-task algorithm comparison (\cref{tab:algo_xtask})
uses a success-rate oracle on T6 but a reward oracle on T3 and T4
out of necessity, so its average should be read as suggestive
rather than a clean cross-oracle aggregate. The FullReval probe is
itself empirical: its $25.0\%$ unsafe rate against the
zero-tolerance gold label on T6 (\cref{tab:safety_split}) is the
$\tau$-tolerance acceptance band, the underlying Monte-Carlo
verdicts are subject to flaky-test-style instability near the
acceptance boundary~\citep{luo2014flaky}, and the size of the band
itself depends on the evaluation resolution
(\cref{tab:resolution}).

\subsection{Internal Validity}
The dominant-skill effect is an associational finding (a single
high-atomic-quality ECM correlates with composition success). We
directly tested and refuted three plausible alternative
mechanisms (hand-off-state coverage, trajectory smoothness, and
trajectory length, \cref{sec:disc}) together with their joint
combination (\cref{sec:app:joint}), and the effect is
verified by an independent re-run with state logging. The
behavioral-distance metrics that would offer a competing prediction
fail on T6, the task on which we collected the off-policy behavioral
data; on
T3 and T4 atomic success collapses to $0\%$ and the ranking carries
no information (\cref{sec:exp:bdist}). Beyond these cross-checks,
the weight-space interpolation of \cref{sec:exp:mechanism} supplies
the mechanistically isolating manipulation that an associational
reading calls for: holding everything fixed except one phase's
weights, composition success follows the atomic probe
point-by-point ($r{=}0.94$) through blended configurations no
training run produced. What remains open is a structural account
of which weight-space property the probe is detecting.

\subsection{External Validity}
The dominant-skill effect underlying the criterion is now
established on two tasks: the dual-arm peg-in-hole task T6 and the
added door-opening task T7 (\cref{sec:exp:door}), the latter also
yielding the critical-path refinement. The headline cost/safety
numbers in \cref{tab:safety_split} and \cref{fig:sens_vs_cost}, the
weight-space interpolation, and the Hybrid Selector benchmark are
nonetheless estimated on the single success-rate-oracled task T6;
T7 is evaluated only at $N{=}30$ and does not repeat the selector
study. The boundary case (T1) and the behavioral-distance
refutation, which is informative only on T6, buttress but do not
multiply that evidence. Four of our six sweep tasks
(T2--T5) reached $0\%$ atomic success under our standard SAC
schedule, so the effect is undefined there. We do not claim it is
absent, only that the schedule yields no measurable candidate
set. We attempted T3\_Stack scaling along two arms before falling
back to the deep-T6 framing. The longer-schedule arm trained for
$26$ iterations on T3\_Stack seed=$2024$ under the default reward
and produced no success-rate transients in any iteration,
indicating that schedule extension alone is not the bottleneck.
The reward-shaping arm modified the environment's reward function
($r_{\text{lift}}$ base $1{\to}2$, alignment bonus $0.5{\to}1$,
$r_{\text{stack}}$ $2{\to}4$) and trained seed=$42$ for $30$
iterations; this arm produced $8$ success-rate transients all at
$1/30$ ($3.33\%$) with reward sustained in $[4, 7]$ versus the
default arm's $[3, 5]$, demonstrating that reward shaping does
move atomic learning, but not enough to clear the $40\%$--$80\%$
Goldilocks zone our paired cross-seed swap matrix requires. T3's
sub-Goldilocks ceiling is therefore robust across both reward
design and schedule extension. All experiments use the same
robosuite Panda arm; cross-embodiment composition stability is an
open direction. The cross-version swap protocol itself simulates
update via independent retraining; realistic continual-learning
updates (fine-tuning, reinforcement learning from human feedback (RLHF), domain adaptation) likely produce
smoother version shifts on which the same atomic-probe criterion
should still apply but with different effect sizes. The dominant
ECM on T6 is identified at $S{=}4$ and should be read as
``the highest-quality ECM'' rather than a strict combinatorial
claim; larger $S$ would likely reveal a continuum of quality with
the same predictive structure.

\subsection{Conclusion Validity}
The statistical conclusions rest on paired McNemar exact-binomial
tests, cluster-permutation tests respecting ECM-level dependence,
$5000$-iteration bootstrap confidence intervals on per-cell rates,
and a $4000$-iteration Monte-Carlo Random-at-matched-cost
baseline. Sample sizes are $48$ paired update events per task,
with every T6 probe evaluated at $N{=}100$ paired episodes; the
$144$-event cross-task aggregate (\cref{sec:app:xtask}) mixes a
success-rate oracle with a reward oracle, retains the original
$N{=}30$ evaluation, and is reported as an appendix extension
rather than primary evidence. The Pareto
frontier in \cref{fig:sens_vs_cost} is a point estimate without
explicit confidence regions on the Hybrid points themselves;
MC error bars are reported for the Random
baseline. Cluster-permutation $p$-values are exact under the
permutation null, and McNemar exact-binomial $p$-values are exact
under independence of paired outcomes; we do not adjust for
multiple comparisons across selectors because the family is small
and each selector is compared against the gold label, not pairwise
against the others. The Hybrid-vs-Random comparison at matched
cost reaches significance on match rate ($P{=}0.039$) and is
directional on sensitivity ($P{=}0.066$); the Hybrid-vs-FullReval
comparison is a weak dominance ($3$ wins, $0$ losses, $p{=}0.25$)
rather than a significant gap, and we phrase it accordingly.
Finally, the dependence of the selector ranking on evaluation
resolution is itself part of the evidence and is reported
explicitly (\cref{tab:resolution}) rather than averaged away.

\section{Conclusion}
\label{sec:conclusion}

We characterized composition stability under skill-update events in
compositional robot policies. On the dual-arm peg-in-hole task a
\emph{dominant-skill effect} governs composition outcomes: a single
high-atomic-quality ECM in the candidate set drives success,
swapping it shifts the rate by up to $52$pp, and a controlled
weight-space interpolation shows composition success following the
blended module's atomic quality point-by-point ($r{=}0.94$); the
effect replicates on a second contact-rich task (Door), where the
dominant module drives the outcome only from an upstream position on
the phase sequence; on a
saturated single-arm pick task the effect is by construction
undefined; and off-policy behavioral-distance metrics fail to
identify the dominant ECM on T6, the task on which they were
measured. On the $48$ success-rate-oracled update events, the
zero-cost atomic probe matches full revalidation with no
detectable difference ($75.0\%$ gold-label
match each) and the Hybrid Selector built on it reaches the best
match on the full-pool gold label ($81.25\%$) at half of
full-revalidation cost, matching or improving on every
alternative's unsafe rate and sensitivity under split-half gold
labels, with
full revalidation's apparent
superiority at coarse evaluation resolution exposed as a
granularity artifact; a $96$-event reward-oracle extension to two
further tasks (\cref{sec:app:xtask}) is directionally consistent
but mixes oracles. This is a principled test-selection criterion specifically targeting
capability-update regression testing in continually-updated skill
libraries. The contribution is demonstrated on two
contact-rich tasks and verified by an independent re-run
with state logging
(\cref{sec:exp:mechanism,sec:exp:door}); broader cross-task and
cross-embodiment generalization
remain the principal open questions, as detailed in \cref{sec:lim}.

\section*{Abbreviations}
\noindent
AUC: area under the curve;\quad
CI: confidence interval;\quad
CPU: central processing unit;\quad
ECM: Embodied Capability Module;\quad
GPU: graphics processing unit;\quad
MC: Monte-Carlo;\quad
ML: machine learning;\quad
PPO: Proximal Policy Optimization;\quad
RL: reinforcement learning;\quad
RLHF: reinforcement learning from human feedback;\quad
RTS: regression test selection;\quad
SAC: Soft Actor-Critic;\quad
SR: success rate.

\section*{Statements and Declarations}

\subsection*{Competing Interests}
The authors declare that they have no known competing financial interests or personal relationships that could have appeared to influence the work reported in this paper.

\subsection*{Funding}
No funding was received for conducting this study.

\subsection*{Ethics Approval}
Not applicable. This study did not involve human participants, human data, or animal subjects. All experiments were conducted in simulation using the robosuite open-source manipulation framework.

\subsection*{Consent to Participate}
Not applicable.

\subsection*{Consent to Publish}
All authors have read and approved the final version of this manuscript and have given their consent for its publication.

\subsection*{Author Contributions}
\textbf{Xue Qin:} Conceptualisation, implementation, experiments, project administration, writing -- original draft.
\textbf{Simin Luan:} Implementation, validation.
\textbf{Cong Yang:} Supervision, conceptualisation, writing -- review and editing.
\textbf{Zhijun Li:} Supervision.

\subsection*{Data Availability}
The evaluation data underlying every table and figure in this manuscript (the JSON outputs for the $N{=}100$ re-evaluation, the weight-space interpolation, the resolution analysis, and the Door replication) are openly available at \url{https://github.com/s20sc/atomic-probe-governance} under the Apache 2.0 license.

\subsection*{Code Availability}
The training, evaluation, statistical-analysis, and figure-generation code is available in the same repository, with the companion simulation framework at \url{https://github.com/s20sc/capability-evolution}.


\bibliography{compositional_related_work}

\end{document}

%% file: related_work.tex

\section{Related Work}
\label{sec:related}

\subsection{Regression Testing and Test Selection}
The problem of deciding which tests to re-run after a change has a
thirty-year tradition in software engineering. The canonical
formulation~\citep{rothermel1996analyzing,rothermel1997safe} keys on
\emph{source-code changes}: a safe regression-test-selection (RTS)
technique identifies the subset of an existing test suite whose
execution can possibly be affected by an edit, using control- and
data-dependence analysis of the source. Empirical studies
established early that the cost/fault-detection trade-off among
selection techniques is context-dependent rather than analytically
decidable~\citep{graves2001empirical}, a conclusion echoed by the
systematic review of the selection
literature~\citep{engstrom2010systematic}. The umbrella survey of
the field~\citep{yoo2012regression} catalogues the minimization,
selection, and prioritization variants and their trade-offs;
within the prioritization variant, the foundational treatment and
the APFD effectiveness measure are due
to~\citet{elbaum2000prioritizing} and~\citet{rothermel2001tcp},
whose cost-cognizant test orderings anticipate the budget-routing
question our Hybrid Selector answers for module updates. RTS has
also moved from research prototype to deployed industrial practice
in continuous integration, where tools such as Ekstazi track
dynamic file-level dependencies per
commit~\citep{gligoric2015ekstazi} and continuous-integration-scale studies report how
selection precision degrades as change granularity
coarsens~\citep{shi2019rtsci}. Learned selectors route continuous-integration budget by
predicted failure probability at industrial
scale~\citep{machalica2019predictive}, and cheap uncertainty
surrogates prioritize test inputs for deep
models~\citep{feng2020deepgini}; both share our margin-gating
intuition (spend the expensive test where a cheap signal is
uncertain) but select over code-triggered test suites or test
inputs rather than over module-update events in a composition.
None of this machinery applies
directly when the
``change'' is a re-trained machine-learned module: the source has not
been edited in a way static analysis can read, and the behavioral
change is what matters, yet it is invisible to dependency graphs.
A separate but adjacent thread on the oracle problem in software
testing~\citep{barr2015oracle} catalogues exactly the difficulty we
face on most of our tasks: a cheap, automatic surrogate for ``did the
system succeed?'' is rarely available. Metamorphic
testing~\citep{chen2018metamorphic} is the canonical response to this
gap, exploiting invariances of the system under test as
pseudo-oracles; \citet{xie2011metamorphicml} applied metamorphic
relations to machine-learned classifiers years before the broader
ML-testing literature took shape. Mutation
testing~\citep{jia2011mutation} evaluates test suites by seeding
artificial faults, and DeepMutation carries the idea to deep
models through weight- and neuron-level
operators~\citep{ma2018deepmutation}; our seed-retraining swap
protocol can be read as a coarse mutation operator over learned
weights whose ``mutants'' are realistic re-trained versions rather
than synthetic corruptions. For learned components specifically, the ML-testing
landscape has grown rapidly~\citep{zhang2022mltesting}, with concrete
exemplars including white-box neural-coverage probing
\citep{pei2017deepxplore} and metamorphic-relation-based testing of
autonomous driving systems~\citep{tian2018deeptest}. All of this work
addresses a \emph{single} learned model under test. Test selection
has likewise been carried into cyber-physical integration
pipelines, including trace-based selection for automotive
continuous integration~\citep{vost2016cirts}, but the selected unit
there remains code under version control rather than a re-trained
learned module. Our setting (the
composition of multiple independently-updatable learned modules,
where the unit under test is the composition and the change is a
component swap) has, to our knowledge, not been studied in the
software-testing literature, and the present paper is positioned to
fill that gap.

\subsection{Typed Composition with Learned Pre/Post-Conditions}
A line of recent work pairs learned skills with explicit symbolic
interfaces. BLADE~\citep{liu2024blade} extracts each high-level
action's pre/post-conditions from language-annotated demonstrations
via an LLM and pairs them with neural controllers; SymSkill
\citep{shao2025symskill} jointly learns predicates, operators, and
skills from unsegmented demonstrations with real-time symbolic
recovery. Generative Skill Chaining~\citep{mishra2023gsc} models the
joint distribution of (precondition, parameters, effect) per skill via
a diffusion model and is the closest neighbor to a stability-aware
view in the present work, although it does not study post-deployment
updates. All such methods study composition under the assumption that
the constituent skills are \emph{static} after construction.

\subsection{Neuro-Symbolic and LLM-Planned Composition}
A complementary thread explicitly synthesizes the symbolic interface
either neuro-symbolically or via an LLM planner. Neuro-Symbolic
Imitation Learning~\citep{neurosymIL2025} discovers PDDL predicates
from demonstrations and refines them with neural skills;
VisualPredicator~\citep{visualpredicator2024} learns neuro-symbolic
predicates for an abstract world model used by a planner.
DeCo~\citep{deco2025} pairs LLM-driven task decomposition with skill
composition for zero-shot long-horizon generalization, and
Text2Motion~\citep{lin2023text2motion} sequences skills through LLM
planning gated by Q-function and geometric feasibility checks. All of
these construct compositions assuming the constituent skills are
fixed; none studies the compositional consequences of updating an
underlying skill, which is the question we ask.

\subsection{Skill Libraries and Skill Chaining}
A second complementary line treats robotic competence as a library of
reusable modules. Voyager~\citep{wang2023voyager}, BOSS
\citep{zhang2023boss}, and LOTUS~\citep{wan2023lotus} are
\emph{append-only}: they grow the library at deployment without
removing or updating existing skills. SayCan~\citep{ahn2022saycan}
and Code-as-Policies~\citep{liang2023cap} pair language-model planning
with primitive skill calls. The most direct neighbours of our
atomic-quality probe sit in the skill-chaining stability thread:
T-STAR~\citep{lee2021tstar} regularizes terminal states at training
time so adjacent skills agree on hand-off distributions; Sequential
Dexterity~\citep{chen2023sequentialdexterity} gates dexterous
policy chaining with a learned transition-feasibility function; and
Value-Informed Skill Chaining~\citep{huang2023visc} gates skill
transitions on a state-value function. All three operate at the
chain-level (preventing bad transitions) rather than the
\emph{update}-level (deciding whether to admit a new candidate
skill into the library), which is the question we ask. None of
this literature formalizes the question of what happens to existing
compositions when one of the constituent skills is later updated.

\subsection{Hierarchical Reinforcement Learning (RL) and Skill Priors}
The methodological foundation of skill modules trained from offline
data and reused for downstream tasks is established by SPiRL
\citep{pertsch2020spirl} and SkiMo~\citep{shi2022skimo}.
DOPPLER~\citep{feng2024doppler} combines options with diffusion
under linear-temporal-logic constraints; LDSC~\citep{ldsc2025} uses
LLM-guided semantic option discovery; LEAGUE~\citep{cheng2023league}
performs guided skill abstraction for long-horizon manipulation;
bottom-up skill discovery from unsegmented
demonstrations~\citep{zhu2022bottomup} is in a similar spirit.
T-STAR~\citep{lee2021tstar} addresses the closely related problem
of terminal-state mismatch between adjacent skills via terminal-state
regularization at training time, while SCaR~\citep{chen2024scar}
regularizes skill chains via dual regularization. Sparse Diffusion
Policy~\citep{wang2024sparsedp} targets continual updates in
diffusion-policy ECMs without forgetting, the closest existing
approach to our deployment scenario.

\subsection{Generalist VLA Policies and the Post-Deployment Update Setting}
Vision--language--action models such as
OpenVLA~\citep{kim2024openvla}, Octo~\citep{octo2024}, $\pi_0$
\citep{black2024pi0}, and RT-2~\citep{brohan2023rt2}, together with
the Open X-Embodiment / RT-X collaboration's large heterogeneous
datasets~\citep{oneill2024openx} and the DROID in-the-wild dataset
\citep{khazatsky2024droid}, are explicitly designed for downstream
fine-tuning, making post-deployment skill updates a routine event.
Recent benchmarks evaluate such generalist policies in simulation
\citep{li2024simplerenv} and in distributed real-world setups
\citep{atreya2025roboarena}, but these evaluate policies as monolithic
units rather than the post-update composition stability we target.
Imitation-learning ECM architectures
\citep{chi2023diffusionpolicy,zhao2023act} are also candidates for
the present protocol but are out of scope here.

\subsection{Compositional Benchmarks}
CompoSuite~\citep{mendez2022composuite} factorizes 256 tasks across
four axes (robot, object, obstacle, objective).
LIBERO~\citep{liu2023libero} and
its robustness extension LIBERO-PRO~\citep{liberopro2025} probe
language-conditioned policies across lifelong-learning suites.
ClevrSkills~\citep{clevrskills2024} provides three explicit levels
of compositional difficulty over ManiSkill2; CALVIN
\citep{mees2022calvin} provides language-conditioned chains of up
to five sub-goals. Across all of these, the unit of generalization
is a \emph{novel} task or composition with the underlying skill
set held fixed.

\subsection{Continual Learning of Skills and Policies}
The continual-learning-for-robotics field is surveyed by
\citet{lesort2019clrobotics}. Module-update testing sits between the
catastrophic-forgetting tradition (elastic-weight consolidation,
\citealt{kirkpatrick2017ewc}) and the incremental-learning taxonomy
of \citet{vandeven2022threetypes}: our cross-version-swap protocol
corresponds most closely to \emph{task-incremental} learning, where
the task identity (the phase) is fixed but the underlying function
(the ECM) is replaced.
An adjacent line of work treats updates not as replacement but as
weight-space \emph{merging}: Model
Soups~\citep{wortsman2022modelsoups} averages weights of fine-tuned
variants, while Task Arithmetic~\citep{ilharco2023taskarithmetic}
edits models via additive task vectors. These approaches keep the
library implicitly versioned in weight space; our protocol applies
symmetrically to either replacement or merging-based updates, since
both ultimately produce a new ECM whose composition stability is
the question of interest. Our work is complementary to all of the
above: we study the effect of single-skill update on
\emph{compositions} that depend on it, rather than on the
single-policy outputs themselves.

\subsection{Off-Policy Policy Selection}
The atomic-quality probe is structurally a per-skill off-policy
evaluator, and the Hybrid Selector is structurally an active
offline-policy-selection procedure: cheap surrogate estimates
warm-start a budgeted online evaluation. Benchmarks for deep
OPE~\citep{fu2021dope} and Active Offline Policy
Selection~\citep{konyushkova2021activeops} establish the
surrogate-vs-online trade-off in the single-policy regime. Our
paired-sampling protocol with bootstrap $95\%$ CIs follows
recommendations from~\citet{agarwal2021precipice} for sparse-trial
RL benchmarks; the McNemar exact-binomial test in
\cref{sec:exp:selector} is supplemented by a cluster-permutation
variant that respects the ECM-level dependence structure of the
$48$ update events. Citation-network analysis of the
typed-composition literature (BLADE, GSC, T-STAR) and the OPE
literature (DOPE~\citep{fu2021dope}, Active
OPS~\citep{konyushkova2021activeops}) finds that, while the two
share foundational RL classics (PPO, Soft Actor-Critic, options), they share no
substantive methodological references and no shared post-2021
citing papers. The Hybrid Selector is structurally an active
offline-policy-selection procedure in the single-policy
regime~\citep{konyushkova2021activeops}, lifted to the
\emph{compositional} regime where the policy unit is a chain of
phase ECMs rather than a single neural network. Bringing OPE rigor
into typed compositional skill libraries is, to our knowledge, an
unaddressed gap that the present work closes for the skill-update
test-selection setting.